\RequirePackage{fix-cm}
\documentclass[smallcondensed]{svjour3}     
\smartqed  
\usepackage{graphicx}
\usepackage[utf8]{inputenc}
\usepackage{graphicx}
\usepackage{todonotes}
\usepackage[colorlinks]{hyperref}
\usepackage{url}
\usepackage{caption}
\usepackage{subcaption}
\usepackage{natbib}
\usepackage{array}
\usepackage{booktabs}
\usepackage{amssymb}
\usepackage{amsmath}
\usepackage{mathtools}
\usepackage{rotating}
\usepackage{mathtools}
\usepackage{longtable}
\usepackage[linesnumbered,ruled,algo2e]{algorithm2e}

\usepackage{soul}
\setstcolor{red}
\newcommand{\ourmethod}{MultiRocket} 
\newcommand{\minirocket}{MiniRocket} 
\newcommand{\rocket}{Rocket} 
\newcommand{\hctwo}{HIVE-COTE 2.0} 
\newcommand{\hivecote}{HIVE-COTE}

\newcommand{\tschief}{TS-CHIEF} 
\newcommand{\inception}{InceptionTime} 
\newcommand{\catch}{catch22} 
 
\newcommand{\ppv}{\textmd{PPV}}
\newcommand{\mmax}{\textmd{Max}}
\newcommand{\lspv}{\textmd{LSPV}}
\newcommand{\mpv}{\textmd{MPV}}
\newcommand{\mipv}{\textmd{MIPV}}

\def\add#1{\textcolor{black}{#1}}

\begin{document}

\title{\ourmethod{}: Multiple pooling operators and transformations for fast and effective time series classification
\thanks{This research has been supported by Australian Research Council grant DP210100072.}
}

\titlerunning{MultiRocket}        

\author{Chang Wei Tan \and
        Angus Dempster \and 
        Christoph Bergmeir \and
        Geoffrey I. Webb
}


\institute{Chang Wei Tan \and Angus Dempster \and Christoph Bergmeir \and Geoffrey I. Webb \at
Department of Data Science and AI\\
Faculty of Information Technology\\
25 Exhibition Walk\\
Monash University, Melbourne\\
VIC 3800, Australia\\
\email{chang.tan@monash.edu,angus.dempster1@monash.edu,christoph.bergmeir@monash.edu,\\
geoff.webb@monash.edu}}

\date{Received: date / Accepted: date}

\maketitle

\begin{abstract}
We propose \ourmethod{}, a fast time series classification (TSC) algorithm that achieves state-of-the-art accuracy with a tiny fraction of the time and without the complex ensembling structure of many state-of-the-art methods.
\ourmethod{} improves on MiniRocket, one of the fastest TSC algorithms to date, by adding multiple pooling operators and transformations to improve the diversity of the features generated.
In addition to processing the raw input series, MultiRocket also applies first order differences to transform the original series. 
Convolutions are applied to both representations, and four pooling operators are applied to the convolution outputs.
When benchmarked using the University of California Riverside TSC benchmark datasets, \ourmethod{} is significantly more accurate than MiniRocket, and competitive with the best ranked current method in terms of accuracy, HIVE-COTE 2.0, while being orders of magnitude faster.
\keywords{Time series classification \and MiniRocket \and Rocket}
\end{abstract}

\section{Introduction}
\label{sec:intro}

Many of the most accurate methods for time series classification (TSC), such as \hctwo{} \citep{middlehurst2021hive}, achieve high classification accuracy at the expense of high computational complexity and limited scalability \citep{middlehurst2021hive}.
Hence scalable TSC has become an important research topic in recent years
\citep{herrmann2021early,tan2020fastee,dempster2021minirocket,dempster2020rocket,shifaz2020ts,lucas2019proximity, schafer2016scalable}. 
{\rocket} and {\minirocket} are the fastest and most scalable among all the proposed scalable TSC methods that achieve state-of-the-art (SOTA) accuracy \citep{dempster2021minirocket,dempster2020rocket}. 
They achieve SOTA accuracy with a fraction of the computational expense of any other method of similar accuracy \citep{dempster2021minirocket,dempster2020rocket}.
Despite their scalability, {\rocket} and {\minirocket} are somewhat less accurate than the variants of {\hivecote} \citep{bagnall2020usage}, including the most recent {\hctwo} \citep{middlehurst2021hive}, which is the current best ranked method with respect to accuracy on 112 datasets in the widely used benchmark UCR archive of time series classification datasets \citep{UCRArchive2018}.

{\minirocket} is built on {\rocket} and is recommended over {\rocket} due to its scalability \citep{dempster2021minirocket}.
We show that it is possible to significantly improve the accuracy of {\minirocket}, with some additional computational expense, by transforming the time series prior to the convolution operations, and by expanding the set of pooling operations used to generate features. 
We call this method {\ourmethod} -- for {\minirocket} with multiple pooling operators and transformations.

{\rocket} and {\minirocket} apply convolutional kernels to the raw input series. The resulting outputs are each summarized by the \emph{Proportion of Positive Values} (\ppv) summary statistic. The resulting values are provided as input features to a simple linear model.
{\minirocket} uses a fixed set of 84 kernels and generates multiple dilations and biases for each kernel, by default producing a total of 10,000 features for the convolution operations. 

\ourmethod{} is based on {\minirocket}, using the same set of kernels as {\minirocket}.
There are two main differences.
First, \ourmethod{} transforms a time series into its first order difference.
Then both the original and the first order difference time series are convolved with the 84 {\minirocket} kernels. 
A different set of dilations and biases is used for each representation because both representations have different lengths (first order difference is shorter by 1) and range of values (bias values are sampled from the convolution output). 
Second, in addition to {\ppv}, \ourmethod{} adds 3 additional pooling operators to increase the diversity and discriminatory power of the extracted features. 
By default, {\ourmethod} produces approximately 50,000 (49,728 to be exact) features per time series (i.e., $6{,}216 \times 2 \times 4$).
For simplicity, when discussing the number of features, we round the number to the nearest 10,000 throughout the paper. 
Finally the transformed features are used to train a linear classifier.

\begin{figure}[t]
    \centering
    \includegraphics[width=\columnwidth]{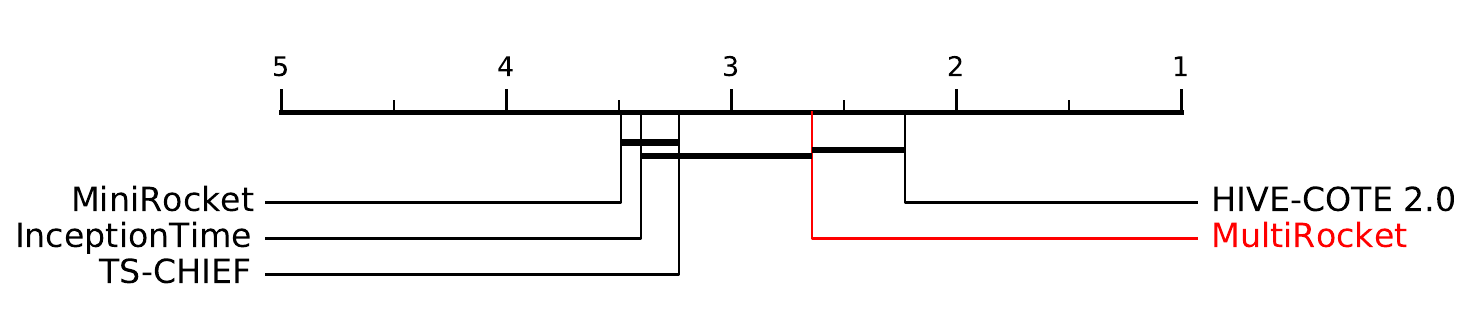}
    \caption{Average rank of \ourmethod{} with the default configuration, in terms of accuracy over 30 resamples of 109 datasets from the UCR archive \citep{UCRArchive2018}, against the top 4 SOTA methods. Classifiers grouped together by a black line (in the same clique) are not significantly different from each other. 
    }
    \label{fig:sota_30}
\end{figure}

Using first order differencing, expanding the set of pooling operators, and increasing the total number of features to 50,000, increases the diversity of the extracted features.
This enhancement makes {\ourmethod} one of the most accurate TSC methods, on average on the datasets in the UCR time series archive \citep{UCRArchive2018}, as illustrated in a critical difference diagram \citep{demvsar2006statistical} shown in Figure \ref{fig:sota_30}.  
Figure \ref{fig:sota_30} shows that {\ourmethod} is significantly more accurate than \minirocket{} (and most top SOTA methods -- see Figure \ref{fig:sota comparison 30 resamples} in our experiments section).
It is also not significantly less accurate than the most accurate TSC method to-date {\hctwo} \citep{middlehurst2021hive}.

\begin{figure}[!t]
	\centering
	\begin{subfigure}{0.49\linewidth}
		\includegraphics[width=\linewidth]{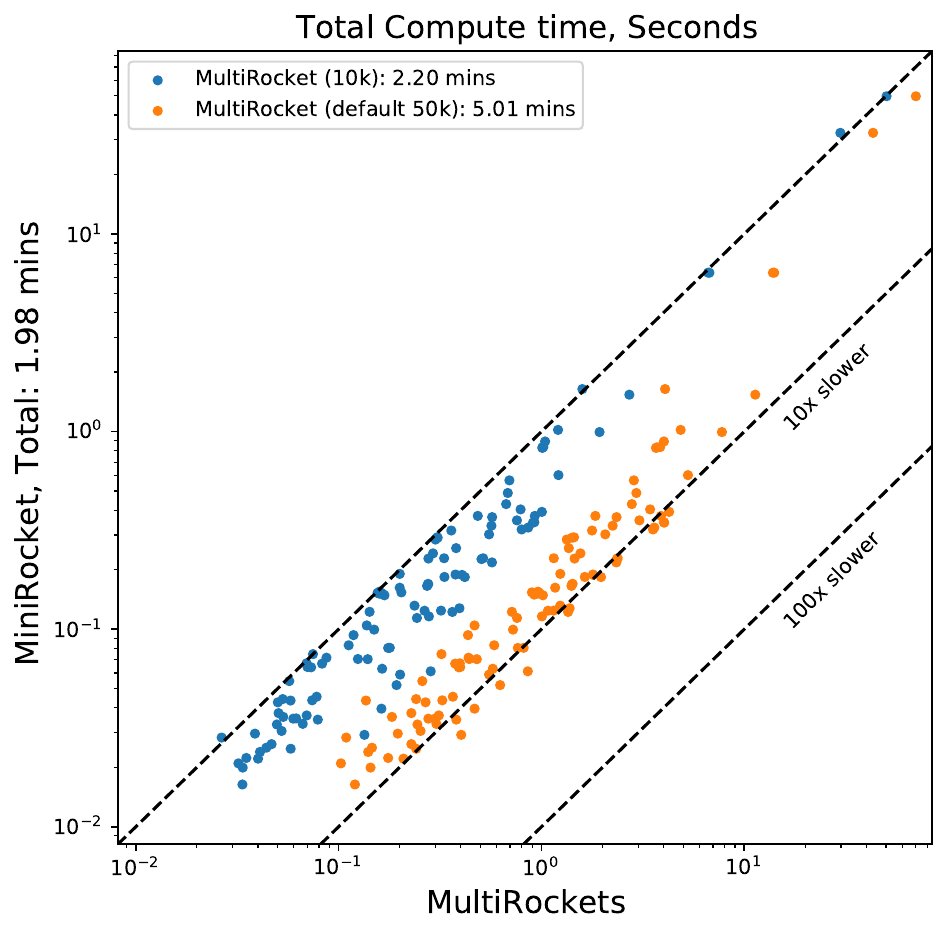}
		\caption{}
		\label{fig:timing}
	\end{subfigure}
	\begin{subfigure}{0.49\linewidth}
		\includegraphics[width=\linewidth]{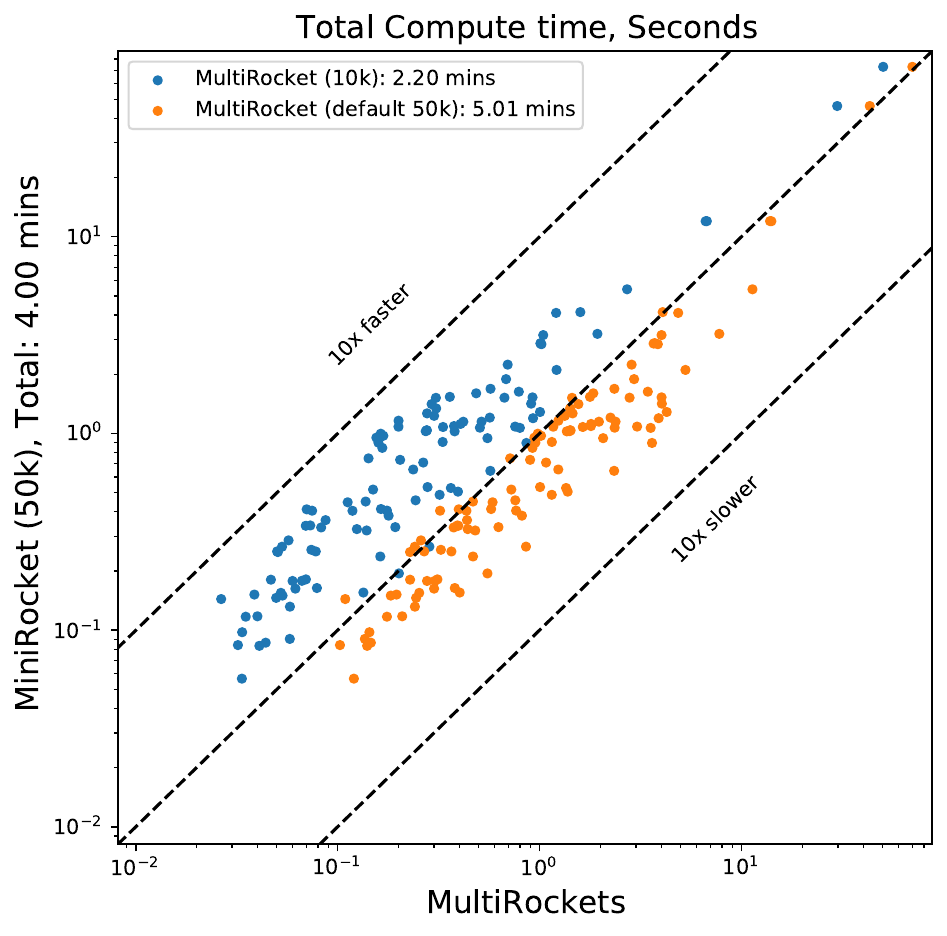}
		\caption{}
		\label{fig:timing 50k}
	\end{subfigure}
	\caption{Total compute time of both \ourmethod{} and \minirocket{} with 10,000 and 50,000 features. Compute times are averaged over 30 resamples of 109 UCR datasets, and run on a cluster using AMD EPYC 7702 CPU with 32 threads. Figure best viewed in color.}
\end{figure}

The use of first order difference transform and additional pooling operators in {\ourmethod} substantially increases the computational expense of the transform over {\minirocket}. 
Figures \ref{fig:timing} and \ref{fig:timing 50k} compare the total compute time (first order difference transform, convolution transforms, training and testing) for {\ourmethod} and {\minirocket}, both with 10,000 and 50,000 features, over 109 datasets from the UCR archive. 
Note that the timings are averages over 30 resamples of each dataset, and run on a cluster using AMD EPYC 7702 CPUs with 32 threads. 
Figure \ref{fig:timing} shows that the default {\ourmethod} with 50,000 features is up to an order of magnitude slower than the default {\minirocket}, which has 10,000 features.
However, the default \ourmethod{} takes only 20\% longer to process the entire repository than \minirocket{} with the same number of features, as illustrated in Figure \ref{fig:timing 50k}. 

Although the default {\ourmethod} (using 50k features) is approximately 10 times slower than the default {\minirocket} (using 10k features), the total compute time for 109 UCR datasets of 5 minutes, using 32 threads, is still orders of magnitude faster than most SOTA TSC algorithms.
The smaller variant of {\ourmethod} with 10,000 features (the same number as the default {\minirocket}) is on average half as fast as {\minirocket} while being significantly more accurate. The relative computational disadvantage of {\ourmethod} relative to {\minirocket} with the same number of features decreases as the number of features increases as the relative impact of once off operations such as taking the derivatives of the series decline as a proportion of total time.


The rest of the paper is organised as follows.  
In Section \ref{sec:related work}, we review the relevant existing work.  
In Section \ref{sec:multirocket}, we describe {\ourmethod} in detail.  
In Section \ref{sec:experiments}, we present our experimental results and conclude our paper.

\section{Related work}
\label{sec:related work}

\subsection{State of the art}
\label{subsec:sota}
The goal of TSC is to learn discriminating patterns that can be used to group time series into predefined categories (classes) \citep{bagnall2017great}.
The accuracy of a TSC algorithm is a measure of its discriminating power.
The current SOTA TSC algorithms with respect to accuracy include \hivecote{} and its variants \citep{middlehurst2021hive,bagnall2020usage,middlehurst2020canonical,middlehurst2020temporal}, \tschief{} \citep{shifaz2020ts}, \minirocket{} \citep{dempster2021minirocket}, \rocket{} \citep{dempster2020rocket} and \inception{} \citep{fawaz2019inceptiontime}.
With some exceptions (namely, {\rocket} and {\minirocket}), most SOTA TSC methods are burdened with high computational complexity.

\inception{} is the most accurate deep learning architecture for TSC \citep{fawaz2019inceptiontime}. 
It is an ensemble of 5 Inception-based convolutional neural networks. 
Ensembling reduces the variance of the model. The resulting method is significantly more accurate compared with other deep learning based TSC methods such as the Fully Convolutional Network (FCN) and Residual Network (ResNet).

\tschief{} was first introduced as a scalable TSC algorithm with accuracy competitive with \hivecote{} \citep{shifaz2020ts}.
It builds on Proximity Forest \citep{lucas2019proximity}, an ensemble of decision trees using distance measures at each node as the splitting criterion.  \tschief{} improves on Proximity Forest by adding interval and spectral based splitting criteria, allowing the ensemble to capture a wider range of representations.

\hivecote{} is a meta-ensemble that consists of the most accurate ensemble classifiers from different time series representation domains \citep{bagnall2020usage,lines2016hive}. 
The original \hivecote{} consists of Ensemble of Elastic Distances (EE) \citep{lines2015time}, Shapelet Transform Classifier (STC) \citep{hills2014classification}, Bag of SFA Symbols (BOSS) Ensemble \citep{schafer2016scalable}, Time Series Forest (TSF) \citep{deng2013time} and Random Interval Forest (RIF) \citep{lines2016hive}, each of them being the most accurate classifier in their respective domains.
The authors showed that \hivecote{} is significantly more accurate than each of its constituent members, and it has stood as a high benchmark for classification accuracy ever since.

Recently {\hctwo} \citep{middlehurst2021hive} was proposed and has been shown to have the best average rank on accuracy against a spread of the SOTA both in the univariate UCR \citep{UCRArchive2018} and the multivariate UEA \citep{bagnall2018uea} time series archives.
{\hctwo}  is a meta-ensemble of four main components, STC, Arsenal, Temporal Dictionary Ensemble (TDE) \citep{middlehurst2020temporal}, and Diverse Representation Canonical Interval Forest (DrCIF) \citep{middlehurst2021hive}. 
{\hctwo} drops EE from the ensemble as EE is not scalable and does not contribute greatly towards the accuracy of {\hivecote} \citep{bagnall2020usage}.
The only module retained from the original {\hivecote} is STC with some additional modifications to make it scalable.
STC in \hctwo{} randomly searches for shapelets within a given contract time and transforms a time series using the distance to each shapelet.
It then employs a rotation forest as the classifier. 
Arsenal is an ensemble of small \rocket{} classifiers with 4,000 features each.
This approach allows the ensemble to return a probability distribution over the classes when making predictions, allowing \rocket{} to be used within the \hivecote{} framework.
The dictionary-based classifier, BOSS was replaced with the more accurate TDE \citep{middlehurst2021hive}.
TDE combines aspects of various earlier dictionary methods and is significantly more accurate than any existing dictionary method \citep{middlehurst2020temporal}.

{\hctwo} updates \hivecote{} by replacing RISE and TSF with DrCIF. 
DrCIF is significantly more accurate than RISE, TSF and its predecessor CIF \citep{middlehurst2020canonical}. 
RISE is an ensemble of various classifiers that derives spectral features (periodogram and auto-regressive terms) from intervals of a time series.
TSF identifies key intervals within the time series, uses simple summary statistics to extract features from these intervals and then applies Random Forests to those features.
DrCIF builds on both RISE and TSF by transforming the time series using the first order difference and periodogram. 
It expands the original set of features used in TSF, using the \catch{} features \citep{lubba2019catch22}.
Diversity is achieved by randomly sampling different intervals and subsets of features for each representation in each tree. 
The use of diverse representations and additional features from the {\catch} feature set within DrCIF results in a considerable improvement in accuracy \citep{middlehurst2021hive}. 
We build on these observations and explore the possibility of extending \minirocket{} with expanded feature sets and diverse representations.

While producing high classification accuracy, most of these methods do not scale well.
The total compute time (training and testing) on the 109 datasets from the UCR time series archive, using a single CPU thread, is around two days for DrCIF, three days for TDE, more than a week for Proximity Forest, and more than two weeks for \hctwo{} \citep{middlehurst2021hive,dempster2021minirocket,middlehurst2020canonical,middlehurst2020temporal}.
On the other hand, \minirocket{} was reported to be able to complete training and testing on 109 datasets within 8 minutes \citep{dempster2021minirocket}.
To be comparable to \ourmethod{}, we ran \minirocket{} on the same hardware, single threaded, and completed the whole 109 datasets just under 4 minutes, while \ourmethod{} with the default 50,000 features takes 40 minutes, an order of magnitude slower (see Figure \ref{fig:timing single thread} in Appendix \ref{app: multi vs mini}).
However, as shown in Figure \ref{fig:timing}, \ourmethod{} was able to complete all 109 datasets in around 5 minutes using 32 threads, while the default \minirocket{} with 10,000 features took around 2 minutes. 
Regardless, \ourmethod{} is still significantly faster than all SOTA methods other than {\minirocket} and highly competitive on accuracy.

\subsection{MiniRocket and Rocket}
\label{subsec:minirocket}

\rocket{} is a significantly more scalable TSC algorithm, matching the accuracy of most SOTA TSC methods \citep{dempster2020rocket}, and taking just 2 hours to train and classify the same 109 UCR datasets using a single CPU core \citep{dempster2021minirocket}. 
{\rocket} transforms the input time series using 10,000 random convolutional kernels (random in terms of their length, weights, bias, dilation, and padding).
It then uses \ppv{} and \mmax{} pooling operators to compute two features from each convolution output, producing 20,000 features per time series.
The transformed features are used to train a linear classifier.
The use of dilation and \ppv{} are the key aspects of \rocket{} in achieving SOTA accuracy.

\minirocket{} is a much faster variant of \rocket{}.
It takes less than 10 minutes to train and classify the same 109 UCR datasets using a single CPU core, while maintaining the same accuracy as \rocket{} \citep{dempster2021minirocket}.
Unlike \rocket{}, \minirocket{} uses a small, fixed set of kernels (with different bias and dilation combinations) and only computes \ppv{} features.
Since \minirocket{} has the same accuracy as \rocket{} and is much faster, \minirocket{} is recommended to be the default variant of \rocket{}  \citep{dempster2021minirocket}. 
In this work, we extend \minirocket{} with first order difference and additional pooling operators to achieve a new SOTA TSC algorithm that is also scalable.
We describe \ourmethod{} in Section \ref{sec:multirocket}.

\subsection{Time series representations}
\label{subsec:representations}

Traditionally, time series analysis involves analysing time series data under different transformations, such as the Fourier transform. 
Different transformations and representations show different information about the time series.
Transforming a time series to a useful representation allows us to better capture meaningful and indicative patterns to discriminate different or group similar time series, thus improving the performance of a model.
A poor representation may lead to lost performance.
For instance, it is easier to analyse time series of different frequencies if they were represented in the frequency domain.
This is known as spectral analysis.
The Fourier transform transforms a time series into the frequency domain, giving a spectrum of frequencies \citep{hannan2009multiple,bracewell1986fourier}.
Then the transformed time series is analysed using the magnitude of each frequency in the spectrum.
A limitation of the Fourier transform is that it only gives information on which frequencies are present but has no information about location and time. 
In consequence, the wavelet transform was proposed to better capture the location of each frequency \citep{vidakovic2009statistical}. 

A recent review \citep{salles2019nonstationary} groups different time series transforms that are often used in time series forecasting tasks into two categories, (1) \emph{mapping} and (2) \emph{splitting} transforms.
Mapping-based transforms map a time series into another representation through a mathematical process such as logarithm, moving average and differencing. 
Splitting-based transforms split a time series into a number of component time series,
such as Fourier and Wavelet transforms that split a time series into different frequencies.
Each component is a simpler time series that can be analysed separately and later be reversed to obtain the original time series representation. 

The derivatives can also be used to capture different information about time series. 
The first order derivative captures the ``velocity'' (rate of change) of the data points in the time series.
The second order derivative then measures the ``acceleration'' of each data point.
\add{\citet{gorecki2013using} combines the original raw series and its first order derivative by weighing the distance of the raw series and the first order derivative series. They showed that their approach achieved better classification accuracy than using the two representations separately.} 
Calculating the exact derivatives of a time series is difficult without knowing the underlying function.
There are many ways to estimate derivatives.
\add{\citet{gorecki2013using} explored 3 different methods and they found that they do not statistically differ from one another.} 
\add{Hence,} we use the simple differencing approach to estimate the derivatives of a time series.

In Section \ref{sec:experiments}, we explore some time series transformation methods to improve the accuracy of \minirocket{} and create \ourmethod{}.

\vspace{-10pt}
\section{MultiRocket}
\label{sec:multirocket}
This section describes \ourmethod{} in detail.
\ourmethod{} shares the overarching architecture of \minirocket{} \citep{dempster2021minirocket} -- it transforms time series using convolutional kernels, computes features from the convolution outputs and trains a linear classifier.
There are two main differences between \ourmethod{} and \minirocket{}.
First is the usage of the first order difference transform and second is the additional 3 pooling operators used per kernel.
The combination of these transformations significantly boosts the classification power of \minirocket{}.
The type of transforms and pooling operators used were tuned on the same 40 ``development'' datasets as used in \citep{dempster2021minirocket,dempster2020rocket} to avoid overfitting the entire UCR archive.  

\vspace{-10pt}
\subsection{Time series representations}
\label{subsec:ts representations}
Diversity is the key to improve a classifier's accuracy. 
\hctwo{} is an accurate TSC classifier because it is a meta-ensemble of a diverse set of time series ensembles, each capturing different representations of a time series, e.g., DrCIF. 

Drawing inspiration from DrCIF, we first inject diversity into \minirocket{} by transforming the original time series into its first order difference.
From this point onward, we refer to the original time series that has not been transformed as the base time series. 
Then convolution is applied to both base and first order difference time series.
We explored different transformation combinations in Section \ref{sec:experiments} and found that this combination works best overall on the 40 ``development'' datasets.
Note that different transformations can be considered depending on the dataset and problem, and we consider this exploration as future work.

The first order difference of a time series 
describes the rate of change of the time series between each unit time step.
This gives additional information about the time series 
, such as identifying the slope of a time series or the presence of certain outliers (or patterns) in a time series that maybe easier to discriminate between two classes.
A given time series $X{=}\{x_1,x_2,...,x_l\}$ is transformed into its first order difference, $X'$  using Equation \ref{eqn:difference}.
We will use this notation to refer to a time series throughout the paper.

\begin{equation}
    X'{=}\{x_t{-}x_{t-1}: \forall t \in \{2,...,l\}\}
    \label{eqn:difference}
\end{equation}

\subsection{Convolutional kernels} 
\label{subsec:kernels}

Now, we describe the convolutional kernels used in \ourmethod{}.
{\ourmethod} uses the same fixed set of kernels as used in {\minirocket} \citep{dempster2021minirocket}, producing high classification accuracy and allowing for a highly optimised transform.  
Note that the enhancement used in \ourmethod{} is also applicable to improve the classification accuracy of \rocket{}.
However, \minirocket{} is preferable over \rocket{} due to its scalability \citep{dempster2021minirocket}.
We refer interested readers to \citep{dempster2020rocket} for details of the kernels used in \rocket{}.
The kernels for \ourmethod{} are characterised in terms of their length, weights, bias, dilation, and padding:

\begin{itemize}
  \item \textbf{Length and weights:} As per {\minirocket}, {\ourmethod} uses kernels of length 9, with weights restricted to two values and, in particular, the subset of such kernels where six weights have the value $-1$, and three weights have the value $2$, e.g., $W {=} [-1, -1, -1, -1, -1, -1, 2, 2, 2]$. This gives a total of 84 fixed kernels.
  \item \textbf{Dilation:} Each kernel uses the same (fixed) set of dilations.  Dilations are set in the range $\{\lfloor 2^{0} \rfloor, ..., \lfloor 2^{\text{max}} \rfloor\}$, with the exponents spread uniformly between 0 and $\text{max} {=} \log_2 ( l_{\text{input}} - 1 ) / ( l_{\text{kernel}} - 1 )$, where $l_{\text{input}}$ is the length of the input time series and $l_{\text{kernel}}$ is kernel length.
  \item \textbf{Bias:} Bias values for each kernel/dilation combination are drawn from the convolution output.  For each kernel/dilation combination, we compute the convolution output for a randomly-selected training example, and take the quantiles of this output as bias values.  (The random selection of training examples is the only random aspect of these kernels.)
  \item \textbf{Padding:} Padding is alternated between kernel/dilation combinations, such that half of the kernel/dilation combinations use padding (standard zero padding), and half do not.
\end{itemize}

\vspace{-10pt}
\subsection{Convolution operation}
\label{subsec:transform}
The base and first order difference time series use different set of dilations and biases to produce the feature maps.
The first order difference time series is shorter by one value than the base time series. 
Hence, the maximum dilation for the first order difference time series will be shorter than the base time series, resulting in a slightly different set of kernels than the base time series.
Additionally, it has a different range of values from the base time series, resulting in a different set of bias values.
Apart from these, the length, weights and padding are the same for both base and first order difference time series.
The convolution operation then involves a sliding dot product between a kernel and a time series.



\vspace{-10pt}
\subsection{Pooling operators}
\label{subsec:features}
After the convolution operations, \ourmethod{} then computes four features per convolution output, $Z$, with length $n$.
These features summarise the values in $Z$ and are also known as pooling operators.
Table \ref{tab:summary of features} shows a summary of the pooling operators used in \ourmethod{}, \emph{Proportion of Positive Values} (\ppv{}), \emph{Mean of Positive Values} (\mpv{}), \emph{Mean of Indices of Positive Values} (\mipv{}) and \emph{Longest Stretch of Positive Values} (\lspv{}). 
The features are illustrated in Figure \ref{fig:feature}.
Algorithm \ref{alg:features} in Appendix \ref{app:features} illustrates the procedure to calculate all four features for a given convolution output, $Z$.

\begin{table}[!t]
    \centering
    \begin{tabular}{c|c|c|c|c}
        \hline
         & \multicolumn{4}{c}{Features} \\
         \hline
         Convolution outputs & \ppv{} & \mpv{} & \mipv{} & \lspv{} \\
         \hline
         $A=[0,0,0,0,0,0,1,1,1,1]$ & 0.4 & 1 & 7.5 & 4 \\
         $B=[1,1,1,1,0,0,0,0,0,0]$ & 0.4 & 1 & 1.5 & 4 \\
         $C=[1,1,0,0,0,0,0,0,1,1]$ & 0.4 & 1 & 4.5 & 2 \\
         $D=[0,0,0,1,1,1,1,0,0,0]$ & 0.4 & 1 & 4.5 & 4 \\
         $E=[0,0,0,0,0,0,10,10,10,10]$ & 0.4 & 10 & 7.5 & 4 \\
         \hline
    \end{tabular}
    \caption{Summary of pooling operators (features) used in \ourmethod{} using a dummy example illustrating different scenarios where \ppv{} will fail to discriminate between different convolution outputs. Each convolution output consists of 6 zeros and 4 positive values giving $\ppv{}=0.4$.}
    \label{tab:summary of features}
\end{table}

\begin{figure}[t]
    \centering
    \includegraphics[width=0.75\columnwidth]{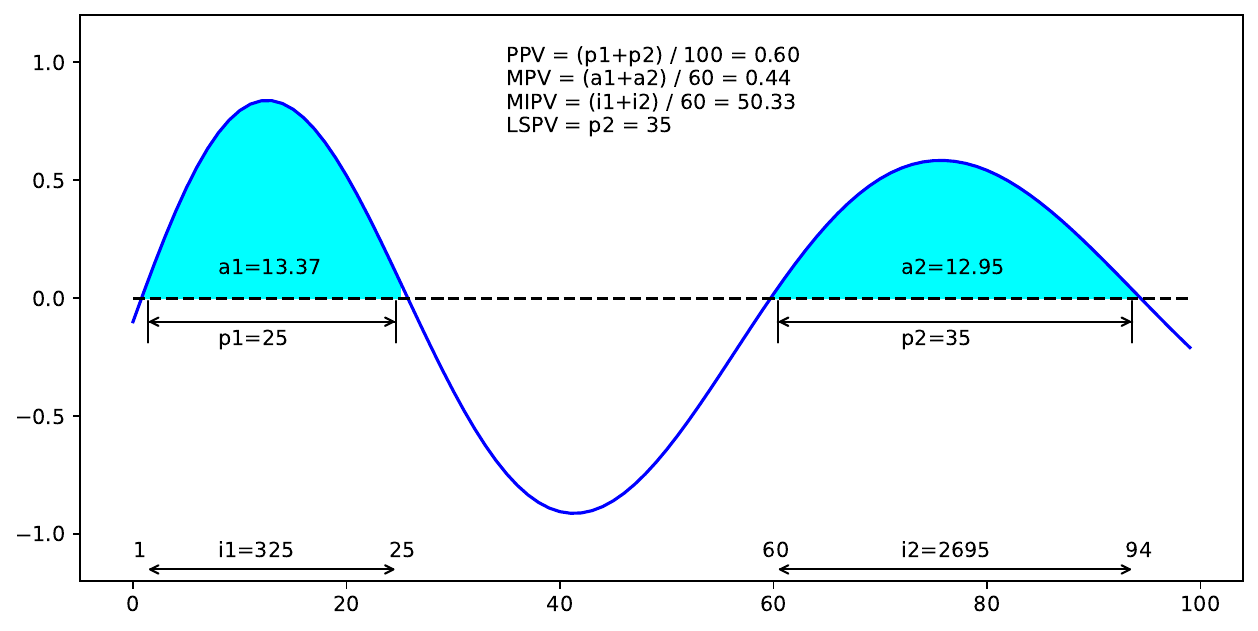}
    \caption{Simple visualisation of the features used in \ourmethod{}, using a convolution output $Z$ of length $n=100$. $p1$ and $p2$ are the number of positive values; $a1$ and $a2$ are the sum of positive values; $i1$ and $i2$ are the sum of the indices of the positive values. Then the features are calculated as shown in the figure and $\lspv{}=p2$ since $p2>p1$.
    }
    \label{fig:feature}
\end{figure}

\vspace{-10pt}
\subsubsection{Proportion of positive values}
\ppv{} was introduced in \rocket{} and was found to be an exceptional feature for \minirocket{}. 
It calculates the \emph{proportion of positive values} from a convolution output $Z$.
\ppv{} is directly related to the bias term which can be seen as a `threshold' for \ppv{}, as described in Equation \ref{eqn:ppv}.
A positive bias value means that \ppv{} is able to capture the proportion of the time series reflecting even weak matches between the input and a given pattern, while a negative bias value means that \ppv{} only captures the proportion of the input reflecting strong matches between the input and the given pattern \citep{dempster2020rocket}.
It is important to note that given \ppv{}, computing the proportion of negative values would not add any extra information as they are complementary to each other. 
Given the exceptional performance and importance of \ppv{} in \minirocket{}, we retain \ppv{} in \ourmethod{}.

\begin{equation}
    \ppv(Z) = \frac{1}{n}\sum_{i=1}^{n}[z_i>0]
    \label{eqn:ppv}
\end{equation}


We augment PPV with three further pooling operators that capture forms of information about the convolutional output to which PPV is blind.

\subsubsection{Mean of positive values}
First, we propose the \emph{Mean of Positive Values} (\mpv{}) to capture the magnitude of the positive values in a convolution output, $Z$ of length $n$, for example, distinguishing A from E in Table \ref{tab:summary of features}.
\mpv{} is calculated using Equation \ref{eqn:mpv} where $Z^{+}$ represents a vector of positive values of length $m$ and $\ppv{}(Z)=|Z^+| / n=m/n$.

\begin{equation}
    \mpv(Z) = \frac{1}{m}\sum_{i=1}^{m}z^{+}_i
    \label{eqn:mpv}
\end{equation}

Similar to \ppv{}, \mpv{} is related to the bias term. 
It captures the intensity of the matches between an input time series and a given pattern -- an information that is available when computing \ppv{} but discarded.
This means that \mpv{} can be computed with negligible additional computational cost.

\subsubsection{Mean of indices of positive values}
The \emph{Mean of Indices of Positive Values} (\mipv{}) captures information about the relative location of positive values in the convolution outputs, for example, distinguishing A from B in Table \ref{tab:summary of features}.
Consider the convolution output $Z$ as an array of values, \mipv{} is computed by first recording the relative location of all positive values in the array, i.e., its indices in the array.
Then the mean of the indices is calculated using Equation \ref{eqn:mipv}, where $I^{+}$ indicates the indices of positive values.
Note that $\ppv{}(Z){=}|I^+|/n{=}m/n$, where $m$ is the number of positive values in $Z$.
In the case where there are no positive values, $m=0$, \mipv{} returns -1 to differentiate from the first index, considering we start with index 0.
For example, the convolution output $A$ in the dummy example in Table \ref{tab:summary of features} has positive values at locations $I^{+}=[6,7,8,9]$ giving $\mipv{}{=}7.5$.

\begin{equation}
    \mipv(Z) = 
    \begin{dcases}
    \frac{1}{m}\sum_{j=1}^{m}i^{+}_j & \text{if } m > 0 \\
    -1 & \text{otherwise}
    \end{dcases}
    \label{eqn:mipv}
\end{equation}

Since $\ppv{}(Z)=|I^+|/n$, the indices of positive values are also available when we are calculating \ppv{}, but currently not used in \minirocket{}. 
Thus, like \mpv{}, \mipv{} can also be computed with negligible additional cost.

\subsubsection{Longest stretch of positive values}
\mipv{} pools all positive values and hence fails to distinguish between many small sequences of successive positive values and a small number of long sequences.
This can provide information of the underlying time series as shown in the example in Appendix \ref{app: lspv}.
%
%
The \emph{Longest Stretch of Positive Values} (\lspv{}) returns the maximum length of any subsequence of positive values in a convolution output, calculated using Equation \ref{eqn:lspv}. 

\begin{equation}
    \lspv(Z) = \max\left[j-i\mid \forall_{i\leq k\leq j}z_k>0\right]
    \label{eqn:lspv}
\end{equation}

This provides a different form of information about the positive values in the convolutional output than is provided by any of the other features, for example, distinguishing C from the remaining series in Table \ref{tab:summary of features}.
Note that calculating \lspv{} comes with a slight overhead over both \mpv{} and \mipv{}.


\subsection{Classifier}
\label{subsec:classifier}
By default, \ourmethod{} produces 50,000 features (49,728 to be exact, using 6,216 kernels, 2 representations and 4 pooling operators).
Like \minirocket{}, the transformed features are used to train a linear classifier.
\ourmethod{} uses a ridge regression classifier by default.
As suggested in \cite{dempster2021minirocket,dempster2020rocket}, a logistic regression classifier is preferable for larger datasets as it is faster to train.
All of our experiments in Section \ref{sec:experiments} were conducted with the ridge classifier. 
The software also supports the logistic regression classifier if required.

\section{Experiments}
\label{sec:experiments}
In this section, we evaluate \ourmethod{} on the datasets in the UCR univariate time series archive \citep{UCRArchive2018}.
We show that \ourmethod{} is significantly more accurate than its predecessor, \minirocket{} and not significantly less accurate than the current most accurate TSC classifier, \hctwo{}.
By default, \ourmethod{} generates $50,000$ features.
We show that even with $50,000$ features, \ourmethod{} is only about 10 times slower than \minirocket{}, but orders of magnitude faster than other current state of the art methods.
Our experiments also show that the smaller variant of \ourmethod{} with $10,000$ features (same number of features as \minirocket{}) is as fast as \minirocket{} while being significantly more accurate. 
Finally, we explore key design choices, including the choice of transformations, features and the number of features.
These design choices are tuned on the 40 ``development'' datasets as used in \citep{dempster2021minirocket, dempster2020rocket} to reduce overfitting of the whole UCR archive.

\ourmethod{} is implemented in Python, compiled via Numba \citep{lam2015numba} and we use the ridge regression classifier from scikit-learn \citep{pedregosa2011scikit}.
Our code and results are all publicly available in the accompanying website, \url{https://github.com/ChangWeiTan/MultiRocket}. 
All of our experiments were conducted on a cluster with AMD EPYC 7702 CPU, 32 threads and 64 GB memory.

\subsection{Comparing with current state of the art}
First, we evaluate \ourmethod{} and compare it with the current most accurate TSC algorithms, namely \hctwo{}, \tschief{}, \inception{}, \minirocket{}, Arsenal, DrCIF, TDE, STC and ProximityForest.
These algorithms\footnote{We obtained the results from \url{https://github.com/angus924/minirocket} for \minirocket{} and  \url{http://www.timeseriesclassification.com/HC2.php} for the rest.} are chosen because they are the most accurate in their respective domains. 
ProximityForest represents the distance-based algorithms;
STC represents shapelet-based algorithms;
While TDE and DrCIF represent dictionary-based and interval-based algorithms respectively.   

For consistency and direct comparability with the SOTA TSC algorithms, we evaluate \ourmethod{} on the same 30 resamples of 109 datasets from the UCR archive as reported and used in \citep{middlehurst2021hive,dempster2021minirocket,bagnall2020usage}.
Note that each resample creates a different distribution for the train and test sets.
Resampling of each dataset is achieved by first mixing the train and test sets, then performing a stratified sampling for train and test sets and maintaining the same number of instances for each resample. 

\begin{figure}[t]
    \centering
    \includegraphics[width=\columnwidth]{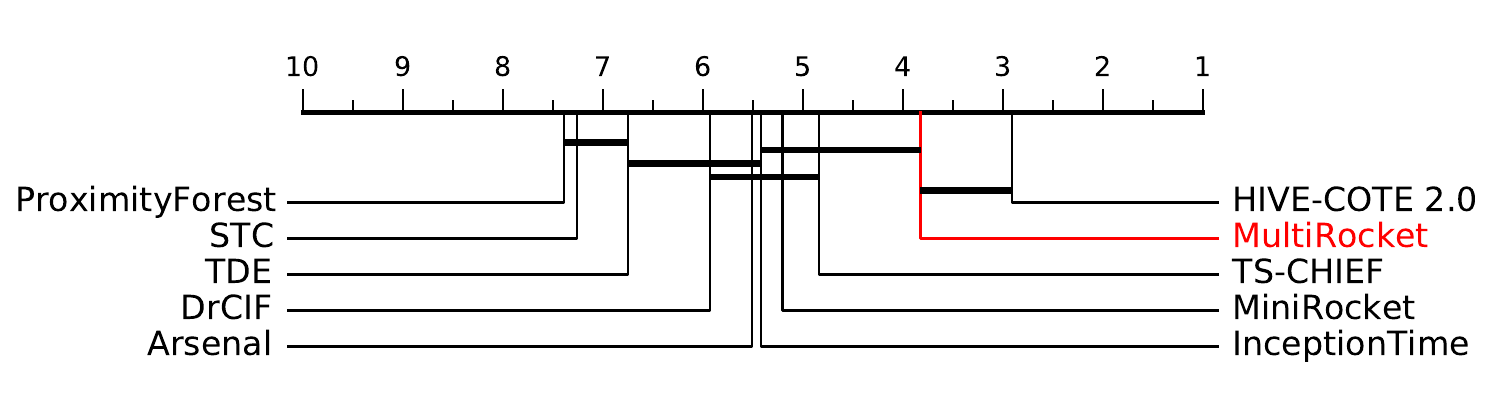}
    \caption{Average rank of \ourmethod{} with the default configuration, in terms of accuracy over 30 resamples of 109 datasets from the UCR archive \citep{UCRArchive2018}, against 9 other SOTA methods. Classifiers grouped together by a black clique indicate that they are not significantly different from each other. 
    }
    \label{fig:sota 30 resamples}
\end{figure}

Figure \ref{fig:sota 30 resamples} shows the average rank of \ourmethod{} against all SOTA methods mentioned. 
The black line groups methods that do not have a pairwise statistical difference using a two-sided Wilcoxon signed-rank test ($\alpha=0.05$) with Holm correction as the post-hoc test to the Friedman test \citep{demvsar2006statistical}.
\ourmethod{} is on average significantly more accurate than most SOTA methods. 
The critical difference diagram with the top 5 algorithms shown in Figure \ref{fig:sota_30} shows that \ourmethod{} is significantly more accurate than its predecessor, \minirocket{} but not significantly less accurate than \hctwo{}, \tschief{} and \inception{}, all of which are ensemble-based algorithms.
Note that \ourmethod{} is one of the few non-ensemble-based algorithms that has achieved SOTA accuracy.
Appendix \ref{app:pairwise comparison} shows the pairwise comparison of some SOTA algorithms.

\begin{figure}[t]
    \centering
    \includegraphics[width=\columnwidth]{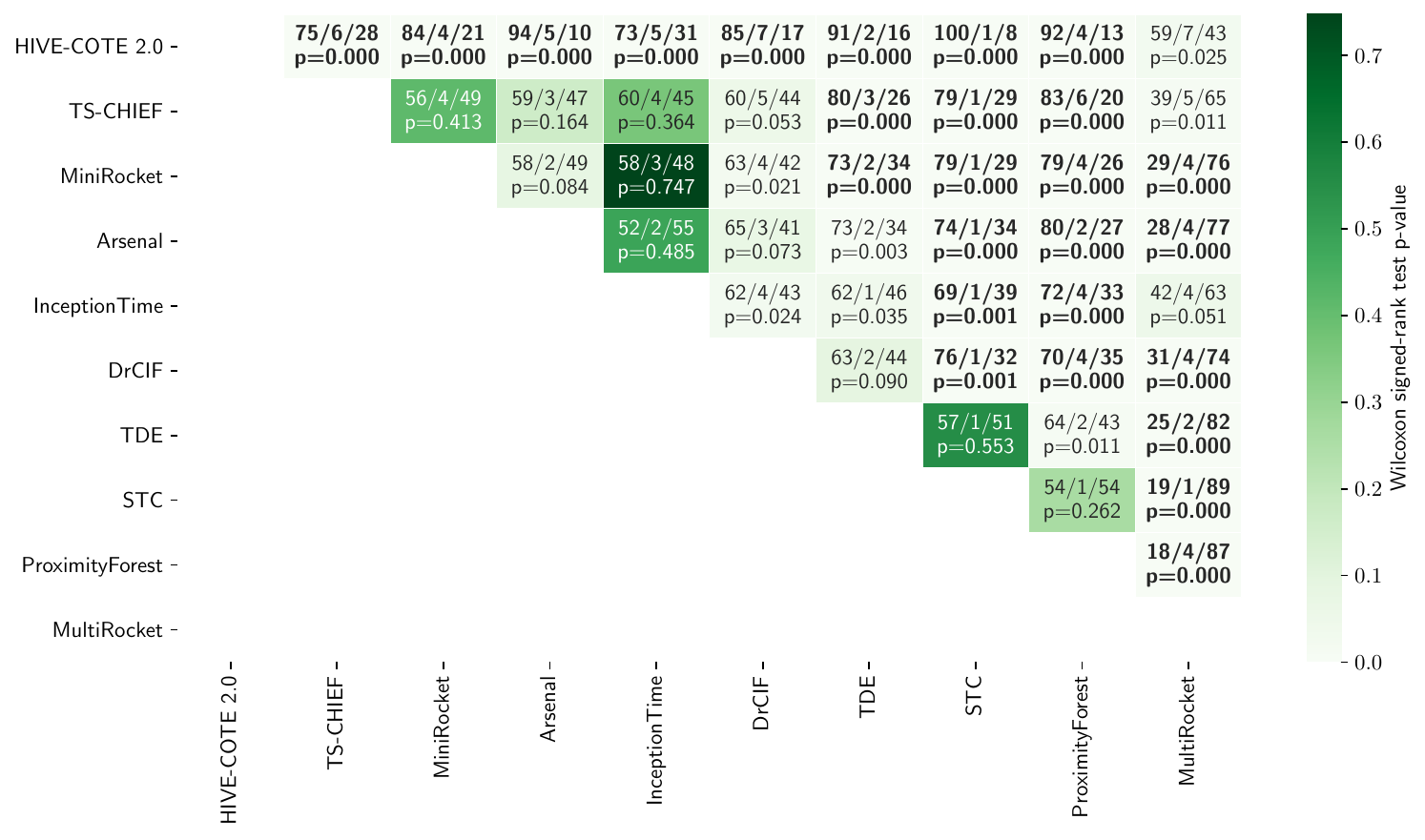}
    \caption{Pairwise statistical significance and comparison of the top SOTA methods. For every cell in the figure, the first row shows the wins/draws/losses of the horizontal method with the vertical method on 30 resamples of the 109 UCR datasets, calculated on the test set; the second row presents the p-value for the statistical significance test, computed using a two-sided Wilcoxon signed rank test. The values in bold indicate that the two methods are significantly different after applying Holm correction.}
    \label{fig:sota comparison 30 resamples}
\end{figure}

Figure \ref{fig:sota comparison 30 resamples} shows the pairwise statistical significance and comparison of the mentioned top SOTA methods.
Every cell in the matrix shows the wins, draws and losses on the first row and the p-value for the two-sided Wilcoxon signed-rank test.
The values in bold indicate that the two methods are significantly different after applying Holm correction.
Overall, as expected and pointed out in \citep{middlehurst2021hive}, \hctwo{} is significantly more accurate than any other method, where the p-values for most of the methods are much less than 0.001, even after applying Holm correction.
\ourmethod{} is the only method with a p-value larger than 0.001 and not significantly different from \hctwo{} after applying Holm correction.
The figure also shows that \ourmethod{} is significantly more accurate than most other methods.

\begin{figure}[!t]
	\centering
	\begin{subfigure}{0.49\linewidth}
		\includegraphics[width=\linewidth]{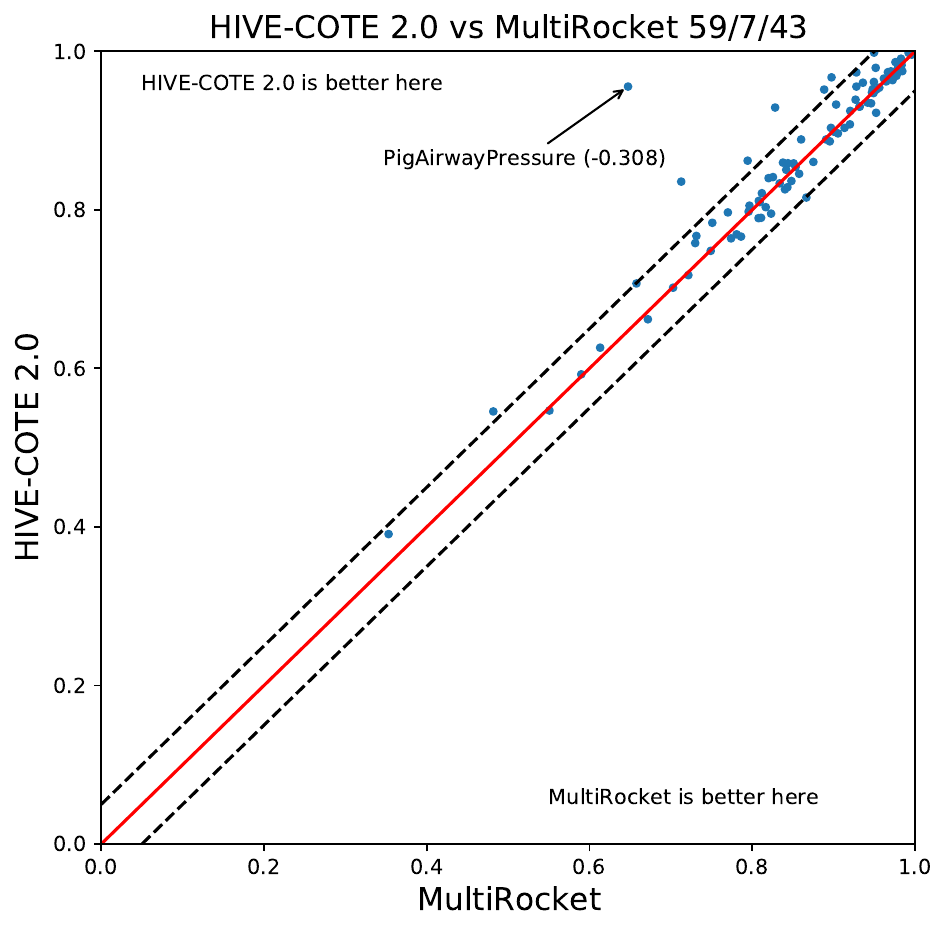}
		\caption{}
		\label{fig:vs hc}
	\end{subfigure}
	\begin{subfigure}{0.49\linewidth}
		\includegraphics[width=\linewidth]{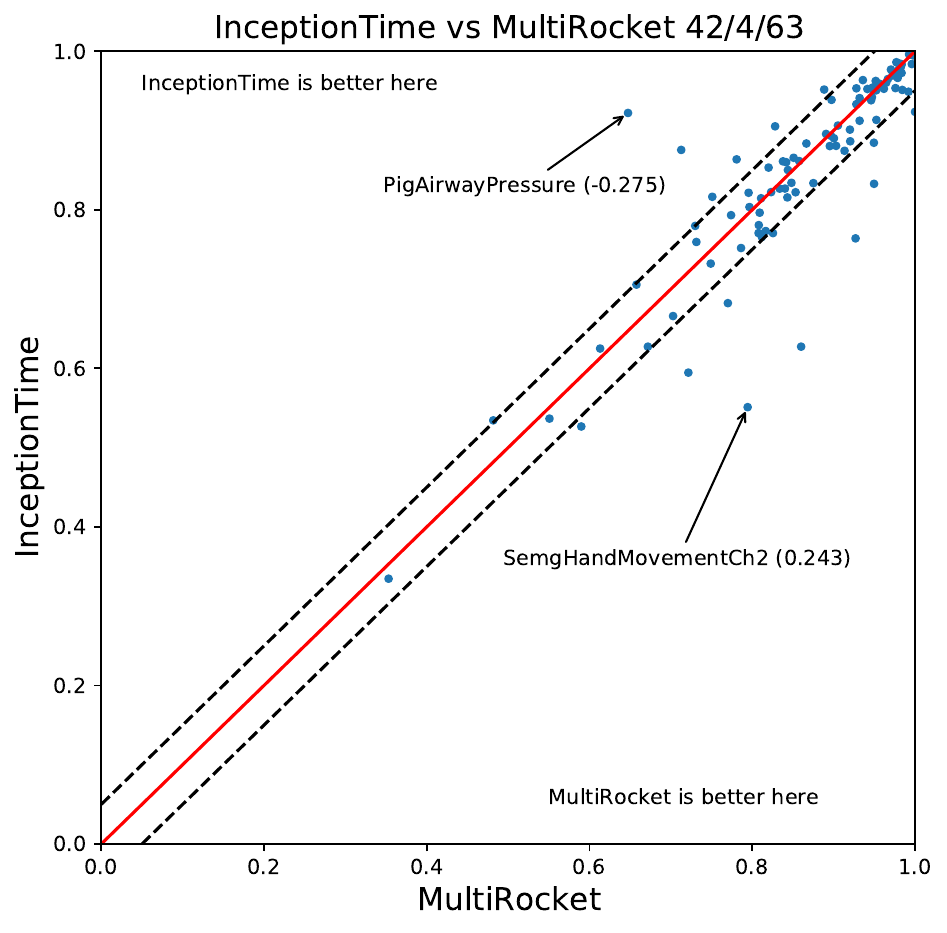}
		\caption{}
		\label{fig:vs inception}
	\end{subfigure}
	\caption{Pairwise accuracy comparison of \ourmethod{} against (a) \hctwo{} and (b) \inception{} on 109 datasets from the UCR archive. Each point represents the average accuracy value over 30 resamples of each dataset. The dotted lines indicate $\pm 5\%$ intervals on the classification accuracy.}
\end{figure}

Although \hctwo{} is significantly more accurate than \ourmethod{} with 59 wins out of 109 datasets, the difference in accuracy between \hctwo{} and \ourmethod{} lies within $\pm5\%$, as shown in Figure \ref{fig:vs hc}, indicating that there is relatively little difference between the two methods.
On the other hand, \ourmethod{} and \inception{} are not significantly different from each other, despite \ourmethod{} having more larger wins, as depicted in Figure \ref{fig:vs inception}.
For instance, \ourmethod{} is most accurate against \inception{} on the \texttt{SemgHandMovementCh2} dataset with accuracy of 0.792 and 0.551. 
While \inception{} is the most accurate against \ourmethod{} on the \texttt{PigAirwayPressure} dataset with accuracy of 0.922 and 0.647.
The large variance in the difference in accuracy between \ourmethod{} and \inception{} implies that both methods are strong in their own ways and that \ourmethod{} can potentially be improved on datasets where \inception{} performed much better. 

\hctwo{}
, \tschief{}
and \inception{} 
are able to capture the different time series representations that have not been able to be captured by \ourmethod{}.
This shows the importance of diversity in classifiers to achieve high classification accuracy.  
However, as shown in Figure \ref{fig:timing}, \ourmethod{} only takes 5 minutes (using 32 threads) to complete training and classification on all 109 datasets, a time that is at least an order of magnitude faster than \hctwo{}, \tschief{} and \inception{}.

As seen on both Figures \ref{fig:vs hc} and \ref{fig:vs inception}, \ourmethod{} performed the worst on the \texttt{PigAirwayPressure} dataset, with the largest difference of 0.308 and 0.275 compared to \hctwo{} and \inception{} respectively.
\rocket{} achieved poor performance on this dataset as pointed out in \citep{dempster2021minirocket} due to the way the bias values are sampled. 
This issue has been mitigated in \minirocket{} by sampling the bias values from the convolution output instead of a uniform distribution, $U(-1,1)$ in \rocket{} \citep{dempster2020rocket}.
\ourmethod{} samples different sets of bias for the base and first order difference series.
It is possible that the first order differences gives rise to the poor performance on this dataset.

\subsection{Runtime analysis}
\begin{figure}[!t]
	\centering
	\begin{subfigure}{0.49\linewidth}
		\includegraphics[width=\linewidth]{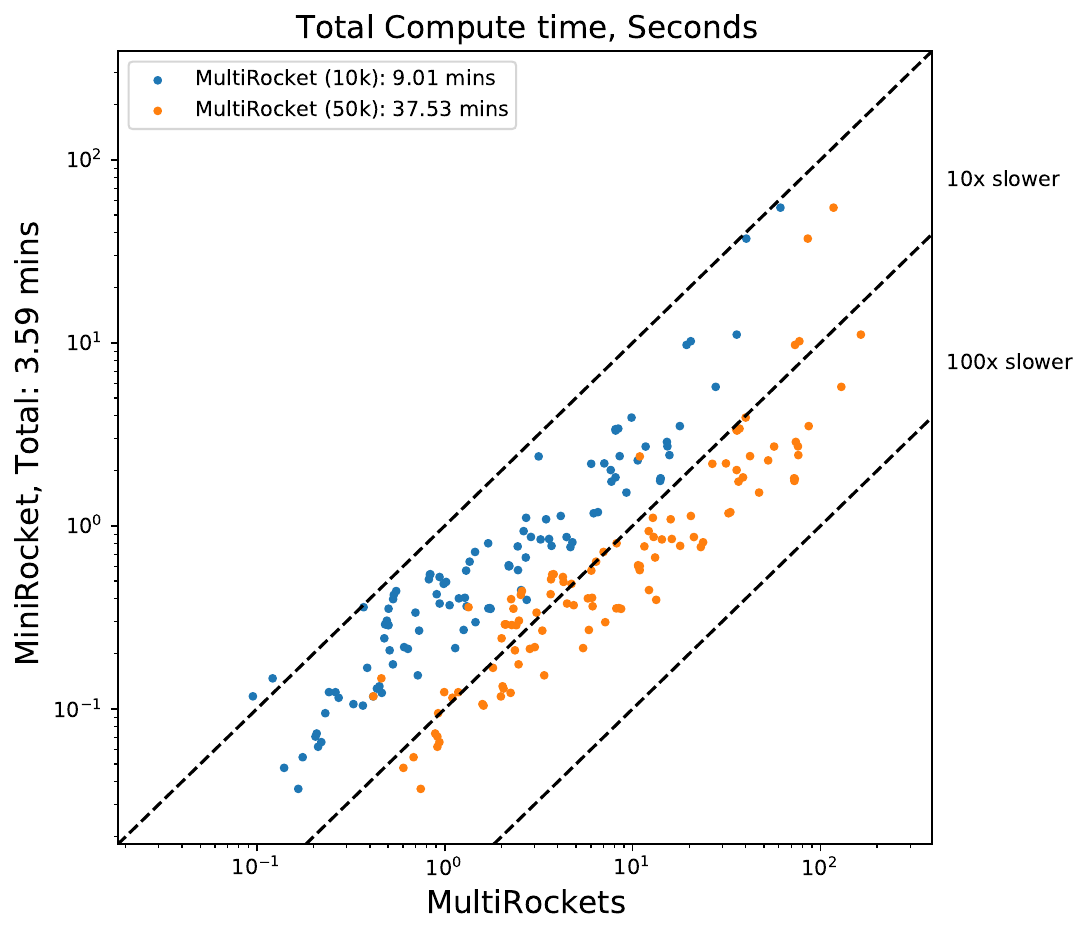}
		\caption{}
		\label{fig:timing single thread}
	\end{subfigure}
	\begin{subfigure}{0.49\linewidth}
		\includegraphics[width=\linewidth]{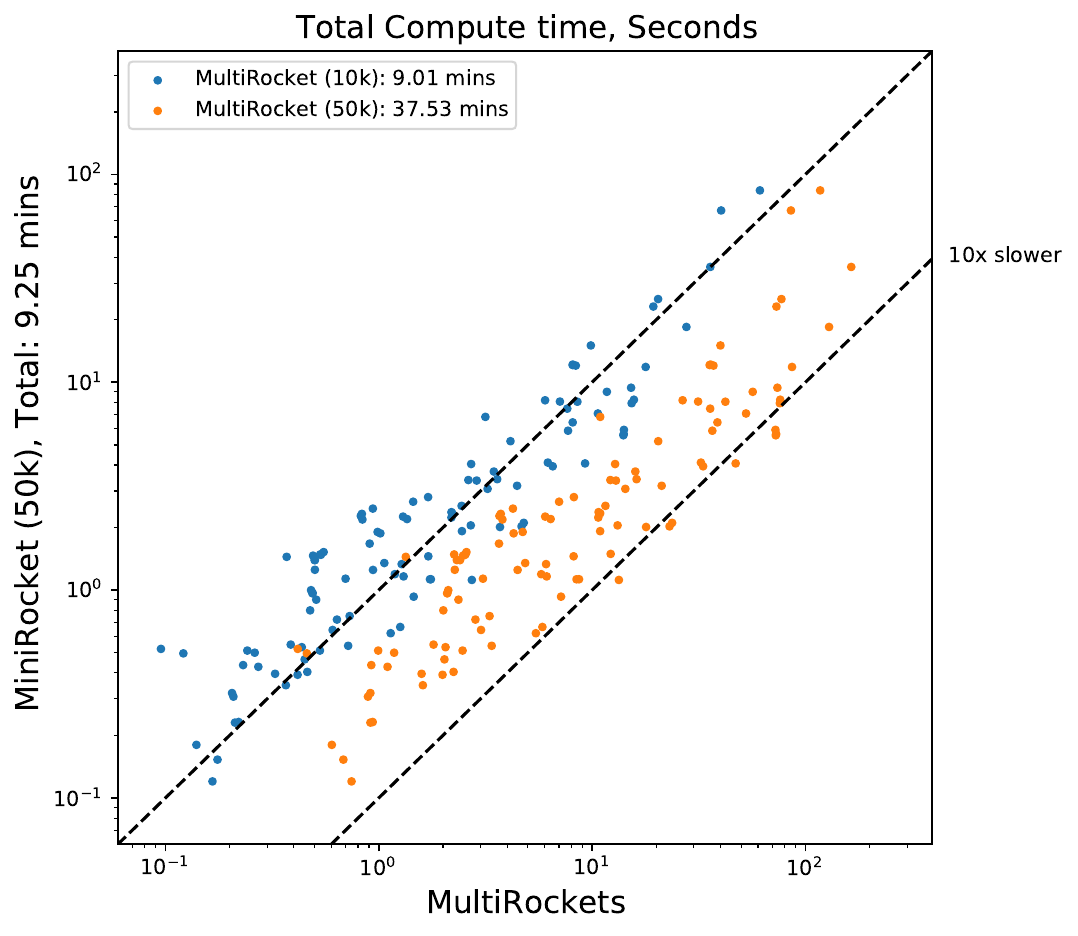}
		\caption{}
		\label{fig:timing single thread 50k}
	\end{subfigure}
	\caption{Total compute time (training and testing) of both \minirocket{} and \ourmethod{}, with 10,000  and 50,000 features. Compute times are obtained from 109 UCR datasets, and run on a cluster using AMD EPYC 7702 CPU with a single thread. Figure best viewed in color.}
\end{figure}

The addition of the first order difference transform and additional 3 features increases the total compute time of \minirocket{}.
Figures \ref{fig:timing single thread} and \ref{fig:timing single thread 50k} show the total compute time (training and testing) of both \ourmethod{} and \minirocket{} with 10,000 and 50,000 features using an AMD EPYC 7702 CPU with a single thread. 
The default \ourmethod{} with 50,000 features is about an order of magnitude slower than the default \minirocket{} with 10,000 features.
Comparing with the same number of 50,000 features, \ourmethod{} is only 4 times slower than \minirocket{}.
This makes sense since \ourmethod{} computes four features per kernel instead of one. 
Taking approximately 40 minutes to complete all 109 datasets, \ourmethod{} is still significantly faster than all other SOTA methods, as shown in Table \ref{tab:runtime}.
However, running \ourmethod{} with 32 threads significantly reduces this time to 5 minutes as shown in Figure \ref{fig:timing}.
Hence it is recommended to use \ourmethod{} in a multi-threaded setting.
Note that \ourmethod{} with 10,000 features is significantly more accurate than \minirocket{} as shown in Appendix \ref{app: multi vs mini}.

All the other SOTA methods have a long run time as reported in \citep{middlehurst2021hive}.
We took the total train time on 112 UCR datasets from \citep{middlehurst2021hive} and show them in Table \ref{tab:runtime} together with a few variants of \ourmethod{} and \minirocket{} with 10,000 and 50,000 features as comparison.
As expected, \minirocket{} is the fastest, taking just under 3 minutes to train.
This is followed by \ourmethod{} that took around 16 minutes.
\rocket{} took approximately 3 hours to train, while Arsenal, an ensemble of \rocket{} took 28 hours. 
The fastest non-Rocket algorithm is DrCIF, taking about 2 days to train, followed by TDE with 3 days.
Finally, the collective ensembles are the slowest taking at least 14 days to train.
Note that the time for \inception{} is not directly comparable as it was trained on a GPU.

\begin{table}[!t]
    \centering
    \begin{tabular}{c|c}
         TSC algorithm & Total train time \\
         \hline
         \minirocket{} (default 10k features) & 2.44 minutes \\
         \ourmethod{} (10k features) & 4.38 minutes \\
         \minirocket{} (50k features) & 5.25 minutes \\
         \ourmethod{} (default 50k features) & 15.77 minutes \\

         \rocket{} & 2.85 hours \\
         Arsenal & 27.91 hours \\ 
         DrCIF & 45.40 hours \\
         TDE & 75.41 hours \\
         \inception{} & 86.58 hours \\
         STC & 115.88 hours \\
         HC2 & 340.21 hours \\
         HC1 & 427.18 hours \\
         TS-CHIEF & 1016.87 hours \\
         \hline
    \end{tabular}
    \caption{Run time to train single resample of 112 UCR problem. \ourmethod{} and \minirocket{} variants are run on a single thread on a cluster using AMD EPYC 7702 CPU with a single thread. The other algorithms are reported in \citep{middlehurst2021hive}.}
    \label{tab:runtime}
\end{table}

\subsection{Ablation study}
So far, we have shown that \ourmethod{} performed well overall.
In this section, we explore the effect of key design choices for \ourmethod{}.
The choices include (A) selecting the time series representations (B) selecting the set of pooling operators and (C) increasing the number of features.

\subsubsection{Time series representations}
\begin{figure}
    \centering
    \includegraphics[width=\columnwidth]{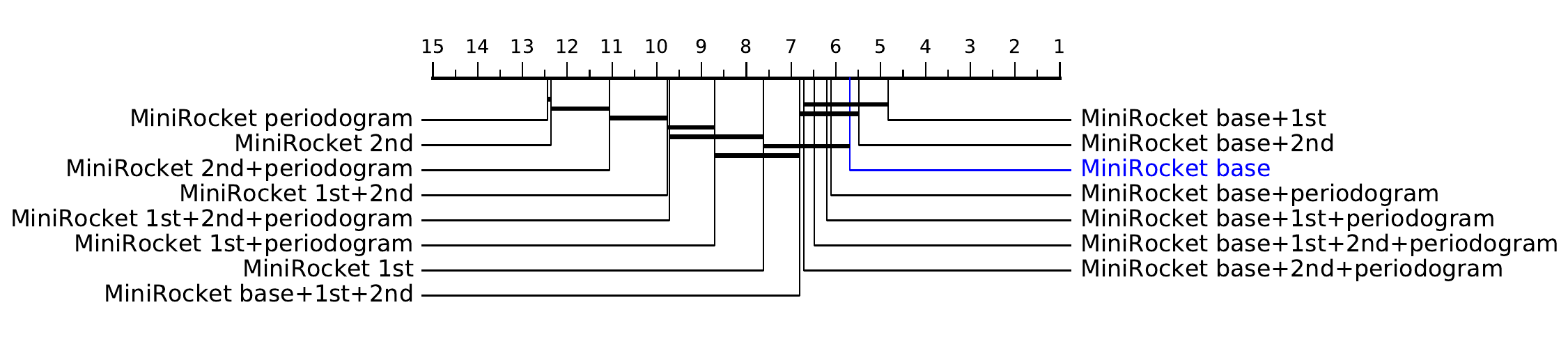}
    \caption{Average rank of different transformations applied on \minirocket{} with 10,000 features.
    }
    \label{fig:minirocket transform}
\end{figure}

We explore the effect of the different representations using \minirocket{} as the baseline.
We consider the first and second order difference to estimate the derivatives of the time series and periodogram to capture information about the frequencies that are present in the time series.
Figure \ref{fig:minirocket transform} shows the comparison of the different combinations of the all 4 representations (including the base time series) for \minirocket{}.
The figure shows that using each of the representation alone does not improve the accuracy, as some information is inevitably lost during the transformation process.
However, combining the base series with either representation improves \minirocket{}, with the first order difference being the most accurate.
The result indicates that adding diversity to \minirocket{} by combining different time series representations with the base time series improves \minirocket{}'s performance.

\begin{figure}
    \centering
    \includegraphics[width=\columnwidth]{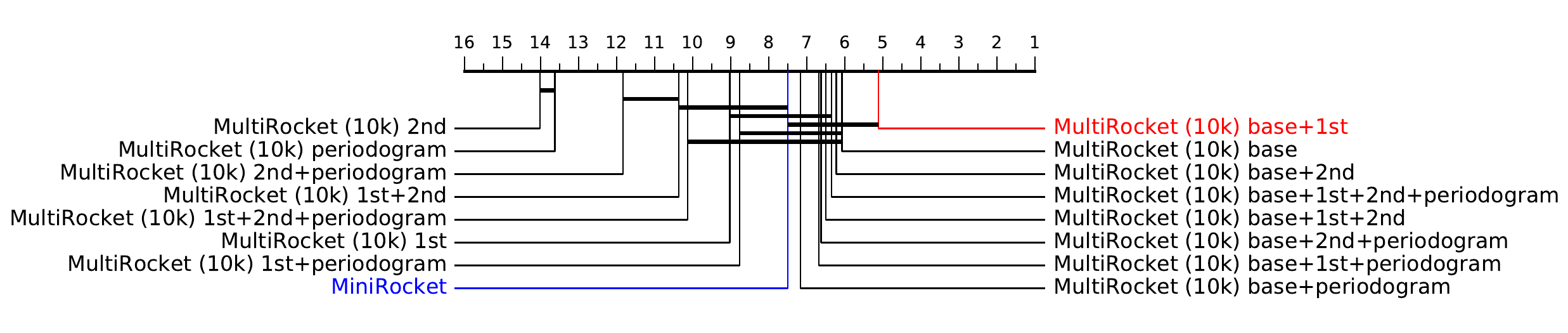}
    \caption{Average rank of different transformations applied on \ourmethod{} with 10,000 features.
    }
    \label{fig:multirocket transform}
\end{figure}

We then perform the same experiment on \ourmethod{} and observed similar results, as shown in Figure \ref{fig:multirocket transform}.
We used the smaller variant of \ourmethod{} to be comparable to \minirocket{}.
In this case, comparing the base versions (\minirocket{} and \ourmethod{} (10k) base) shows that adding the additional 3 pooling operators also improves the discriminating power of \minirocket{}, as indicated in the discussion in Appendix \ref{app: multi vs mini}.

\subsubsection{Pooling operators}
\begin{figure}
    \centering
    \includegraphics[width=\columnwidth]{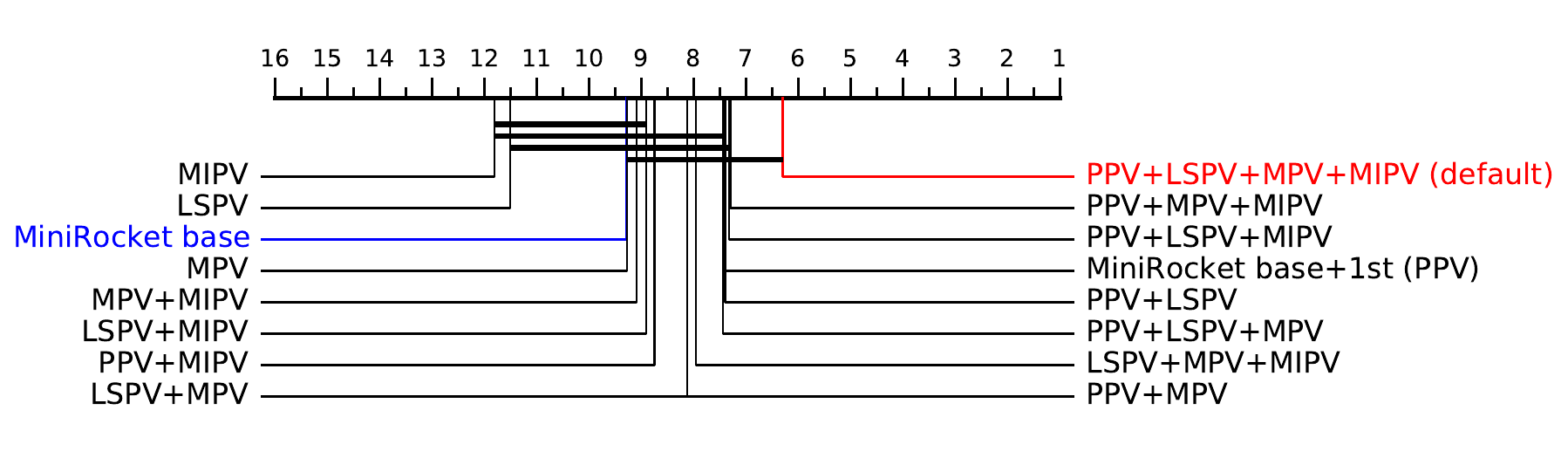}
    \caption{Average rank of different feature combinations applied to base and first order difference with 10,000 features.
    }
    \label{fig:10k features}
\end{figure}

The previous section shows that applying convolutions to the base and first order difference series improves the discriminating power of \minirocket{} and \ourmethod{}.
Hence it is chosen as the default for \ourmethod{}.
Now, we explore the effect of different combinations of pooling operators used by each kernel on classification accuracy.
Figure \ref{fig:10k features} compares the different pooling operator combinations of \ourmethod{} with 10,000 features with the baseline \minirocket{} and \minirocket{} with base and first order difference series.
The result shows that the variant using all pooling operators performed the best overall. 
This confirms our justification of using all four pooling operators in Section \ref{sec:multirocket}.
Figure \ref{fig:10k features} also shows that \ppv{} is a strong feature, where most of the combinations did not perform better than using \ppv{} alone. 
The use of each pooling operator alone (without the combination) also performed significantly worse than \ppv{}.

\subsubsection{Number of features}
\begin{figure}
    \centering
    \includegraphics[width=\columnwidth]{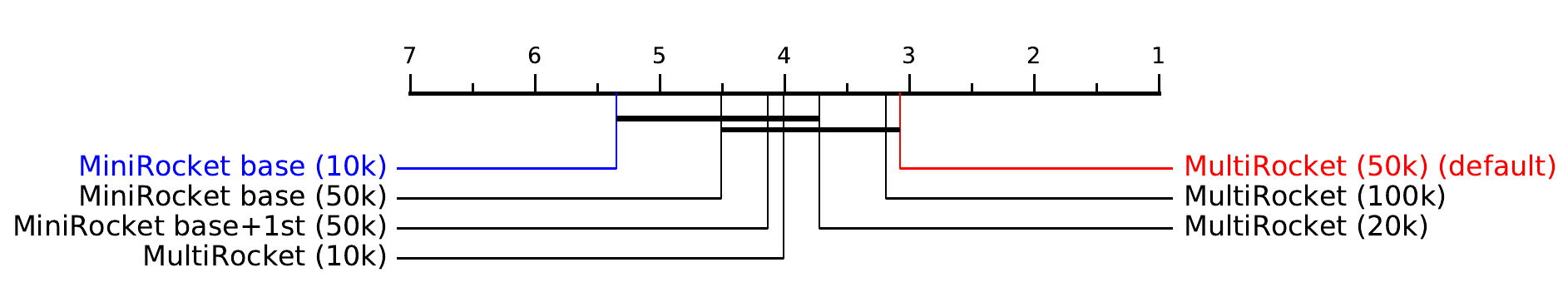}
    \caption{Average rank of increasing number of features on the base and first order difference time series.
    }
    \label{fig:base 1st features}
\end{figure}

\begin{figure}[!t]
	\centering
	\begin{subfigure}{0.49\linewidth}
		\includegraphics[width=\linewidth]{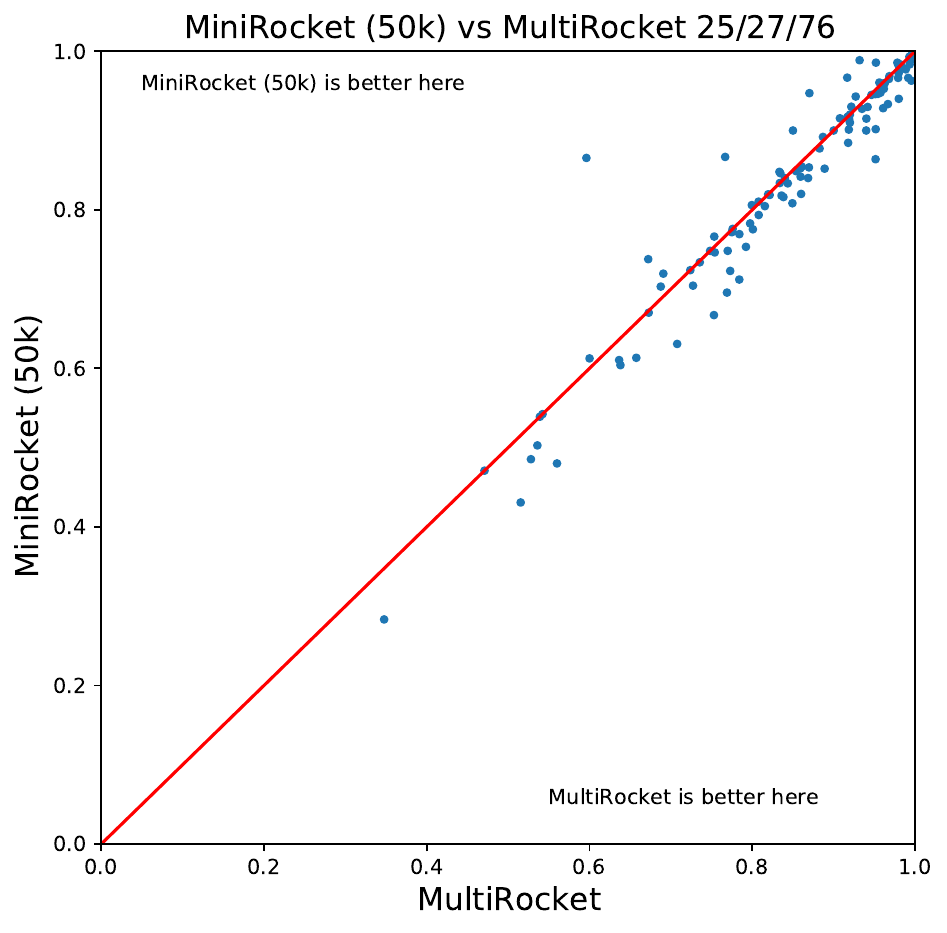}
		\caption{}
		\label{fig:vs mini 50}
	\end{subfigure}
	\begin{subfigure}{0.49\linewidth}
		\includegraphics[width=\linewidth]{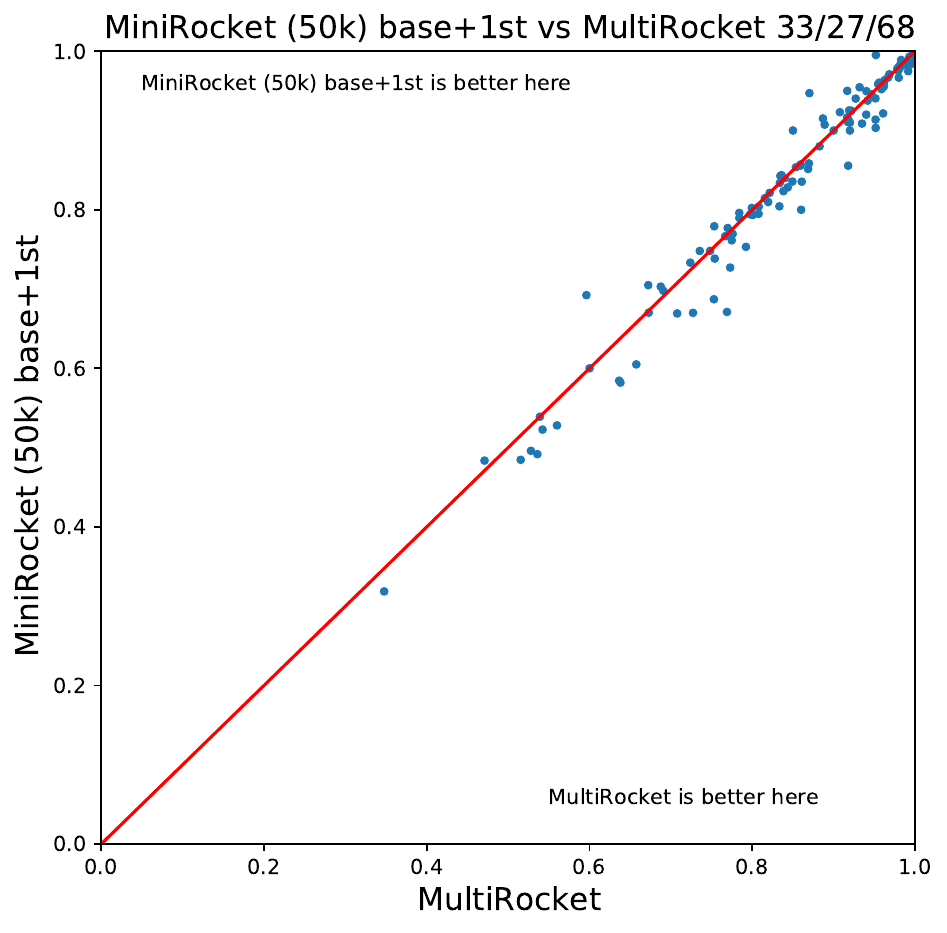}
		\caption{}
		\label{fig:vs mini 50 base+1st}
	\end{subfigure}
	\caption{Pairwise accuracy comparison of \ourmethod{} against (a) \minirocket{} with 50,000 features and (b) \minirocket{} with 50,000 features, base and first order difference on 109 datasets from the UCR archive. Each point represents the average accuracy value over 30 resamples of the each dataset}
\end{figure}

The default setting of \ourmethod{} uses the combination of the base and first order difference and extracts 4 features per convolution kernel.
In this section, we explore the effect of increasing the number of features in \ourmethod{}.
Figure \ref{fig:base 1st features} shows the comparison of \ourmethod{} with different numbers of features.
We also compare with the default \minirocket{}, \minirocket{} with 50,000 features and \minirocket{} with base and first order difference.
Overall, using 50,000 features is the most accurate and there is little benefit in using 100,000 features as more and more features will be similar to one another.
A similar phenomenon was shown in \cite{dempster2021minirocket}.
Figures \ref{fig:vs mini 50} and \ref{fig:vs mini 50 base+1st} show that \ourmethod{} with 50,000 features is significantly more accurate than both \minirocket{} with 50,000 features and with the first order difference.
\ourmethod{} is more accurate on 76 and 68 datasets respectively.
The results show that the increase in accuracy is not just due to the large number of features but also due to the diversity in the extracted features using the four pooling operators and first order difference.
Therefore \ourmethod{} uses 50,000 features by default.

\section{Conclusion}
\label{sec:conclusion}
We introduce \ourmethod{}, by adding multiple pooling operators and transformations to \minirocket{} to improve the diversity of the features generated.
\ourmethod{} is significantly more accurate than \minirocket{} but not significantly less accurate than the most accurate univariate TSC algorithm, \hctwo{} on the UCR archive. 
While being approximately 10 times slower than \minirocket{}, \ourmethod{} is still significantly
faster than all other state-of-the-art time series classification algorithms.

\ourmethod{} applies first order differencing to transform the time series. 
Then four pooling operators \ppv{}, \mpv{}, \mipv{} and \lspv{} are used to extract summary statistics from the convolution outputs of the base and first difference series. 
As the application of convolutions to time series is designed to highlight useful properties of the series, it seems likely that further development of methods to isolate the relevant signals in these convolutions will be highly
productive.
Besides, different transformation methods can also be explored to further improve the diversity of \ourmethod{}.
Further promising future directions include exploring the utility of \ourmethod{} on multivariate time series, regression tasks \citep{tan2021time} and beyond time series data.


\begin{acknowledgements}
We would like to thank Professor Eamonn Keogh, Professor Tony Bagnall and their team who have provided the UCR time series classification archive \citep{UCRArchive2018} and making a comprehensive benchmark results widely available. Finally, we thank Hassan Fawaz for providing open source code to draw critical difference diagrams.
\end{acknowledgements}


%
%

\bibliographystyle{spbasic}      
\bibliography{biblio}   

\newpage
\appendix
\section{Features in \ourmethod{}}
\label{app:features}
Algorithm \ref{alg:features} illustrates the procedure to calculate all four features for a given convolution output, $Z$.
First, we initialise the variables in lines 1 to 5, such as a counter for positive values, $p$; $\mu$ to calculate the mean values; $i$ for the mean of indices; and two variables (last\_val and max\_stretch) to remember the longest stretch of positive values. 
Lines 6 to 16 iterate through $Z$ and extract the required information to compute the features.
After iterating through $Z$, we do a final check on the longest stretch in lines 17 to 19.
Finally the features are computed in lines 20 to 24 and the algorithm terminates on line 25 by returning the feature vector. 

\begin{algorithm2e}
    \caption{\textsc{ComputeFeatures}($Z$)}
    \label{alg:features}
    \DontPrintSemicolon
    
    \SetKwFunction{length}{length}
    \SetKw{KwTo}{to}
  
    \KwIn{$Z$: A convolution output after applying the kernels}
    \KwResult{$F$: An array of 4 features, $(\ppv{},\mpv{},\mipv{},\lspv{})$}
    
    \BlankLine
    \tcp{initialise}
    $p \gets 0$ \tcp*{positive count}
    $\mu \gets 0$ \tcp*{mean value}
    $i \gets 0$ \tcp*{mean of indices}
    $\text{last\_val} \gets 0$ \tcp*{last non-positive value}
    $\text{max\_stretch} \gets 0$ \tcp*{longest stretch so far}
    \For{$j \gets 0$ \KwTo $Z.\length-1$}{
        \uIf{$Z_j > 0$}{
        $p \gets p + 1$ \;
        $\mu \gets \mu + Z_j$ \;
        $i \gets i + j$ \;
        } 
        \uElse{
            \If{$(j - \text{last\_val}) > \text{max\_stretch}$}{
                $\text{max\_stretch} \gets j - \text{last\_val}$ \;
            }
            $\text{last\_val} \gets j$ \;
        }
        
    }
    \tcp{check the last value of $Z$}
    \If{$(Z.\length - 1 - \text{last\_val}) > \text{max\_stretch}$}{
        $\text{max\_stretch} \gets Z.\length - 1 - \text{last\_val}$ \;
    }
    Let $F$ be an array of 4 \; 
    $F_0 \gets p / Z.\length$ \tcp*{calculate \ppv{}}
    $F_1 \gets \mu / p$ \tcp*{calculate \mpv{}}
    $F_2 \gets i / p$ \tcp*{calculate \mipv{}}
    $F_3 \gets \text{max\_stretch}$ \tcp*{calculate \lspv{}}
    \Return $F$
\end{algorithm2e}

\newpage
\section{Example for Longest Stretch of Positive Values}
\label{app: lspv}
\begin{figure}[h]
    \centering
    \includegraphics[width=\columnwidth]{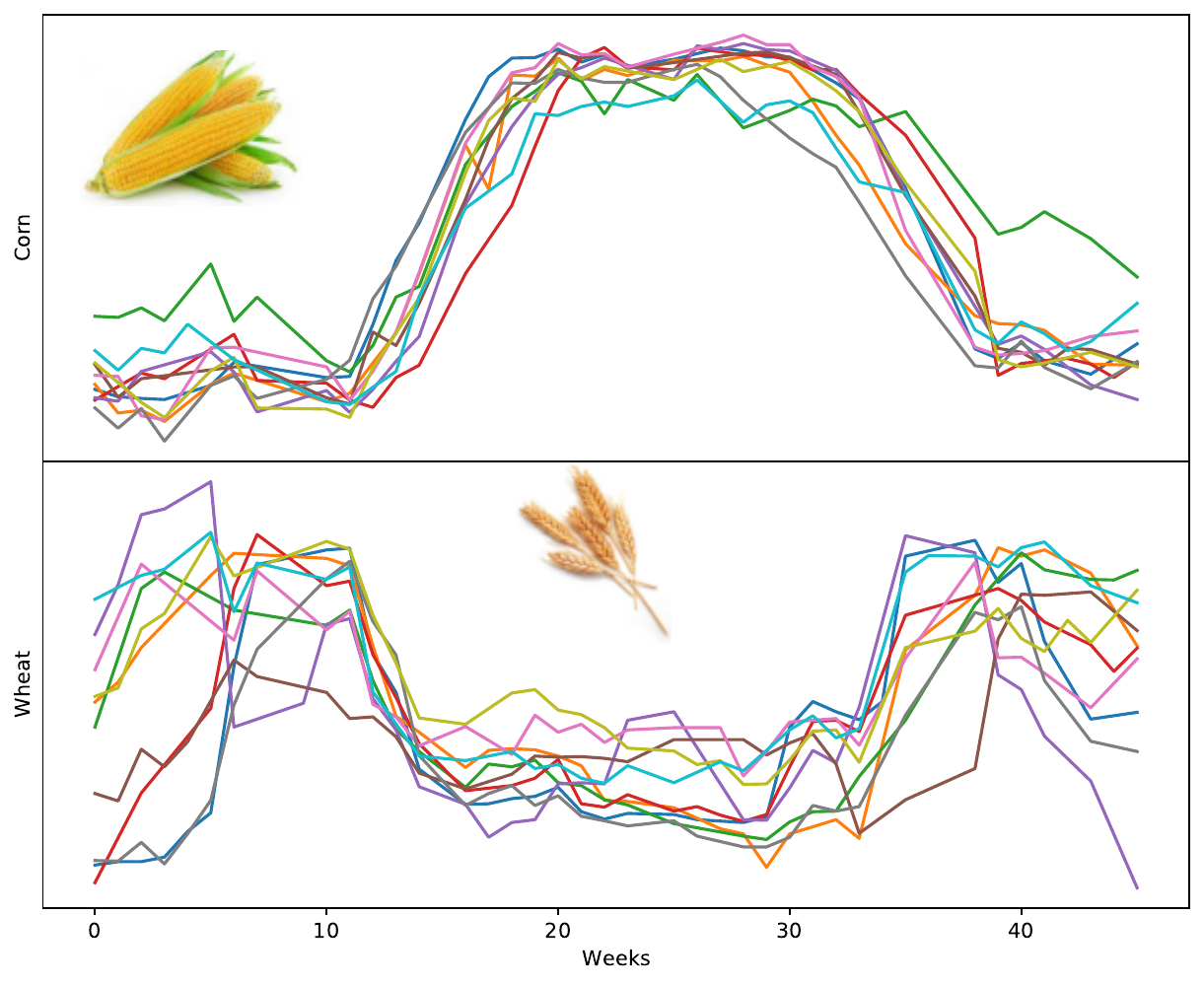}
    \caption{Illustration of convolution outputs $C$ and $D$ from Table \ref{tab:summary of features}, with the example of differentiating corn and wheat (in southern France) using the satellite image time series, obtained from \citep{tan2017indexing}. Corn is a summer crop having peaks in the middle of the year ($D$); while wheat is a winter crop, having peaks at the start and end of the year ($C$).}
    \label{fig:lspv example}
\end{figure}

The \emph{Mean of Indices of Positive Values} (\mipv{}) has a limitation in differentiating convolution outputs with positive values at the start and end of convolution outputs with positive values in the middle. 
\mipv{} would give the same 4.5 for both  convolution outputs $C$ and $D$ in Table \ref{tab:summary of features}.
However, it is obvious that both $C$ and $D$ come from different time series that could potentially be from two different classes.
The underlying time series for $C$ has patterns appearing at the start and end of the time series while $D$ has the same pattern appearing in the middle of the time series. 
For example, differentiating summer crops from winter crops (based on their satellite image time series), where summer crops could have peaks in the middle of the year for three months (Jun-Sep), while winter crops have peaks at the start and end (Dec-Feb), also three months. 
Figure \ref{fig:lspv example} illustrates the example of \lspv{}, using the satellite image time series of corn and wheat, taken in southern France \citep{tan2017indexing}.
Therefore, we propose the \emph{Longest Stretch of Positive Values} (\lspv{}) to mitigate this issue.

\newpage
\section{Pairwise comparisons}
\label{app:pairwise comparison}
In this section, we show the pairwise comparison of SOTA methods against \ourmethod{}.
The figures show that \ourmethod{} is significantly more accurate than all of them, although most of the improvements are within $\pm 5\%$.

\begin{figure}[!h]
	\centering
	\begin{subfigure}{0.49\linewidth}
		\includegraphics[width=\linewidth]{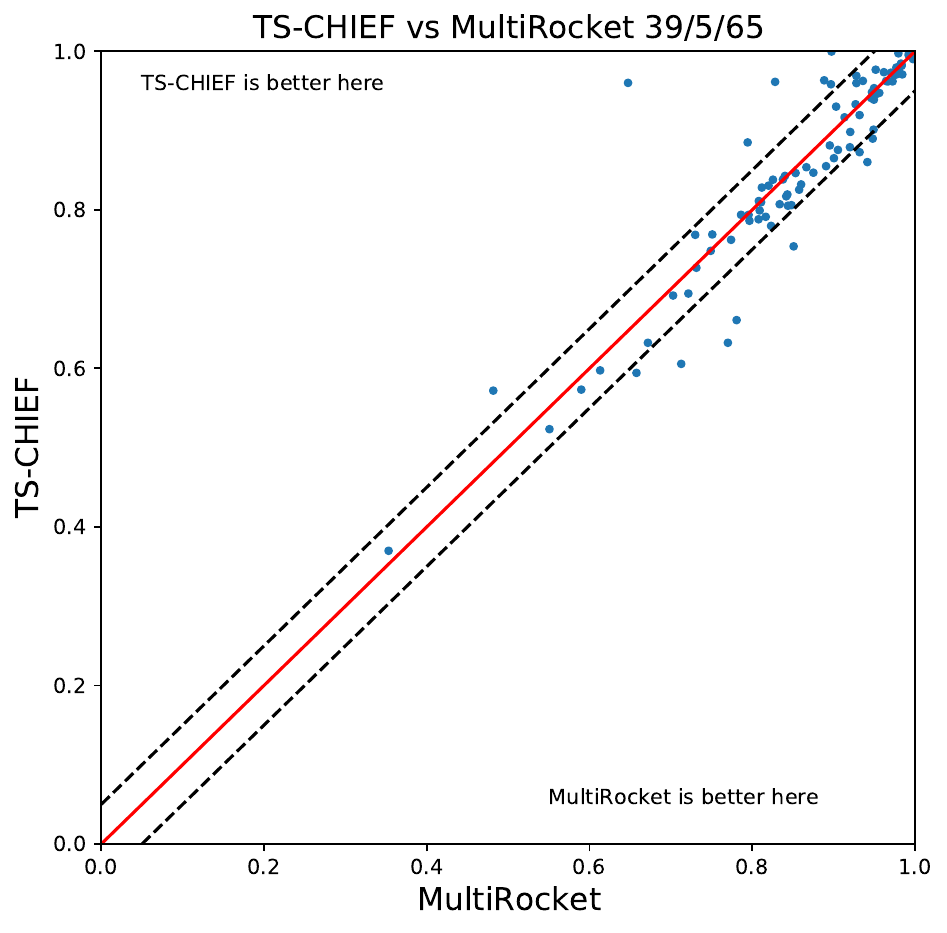}
		\caption{}
		\label{fig:tschief vs multi}
	\end{subfigure}
	\begin{subfigure}{0.49\linewidth}
		\includegraphics[width=\linewidth]{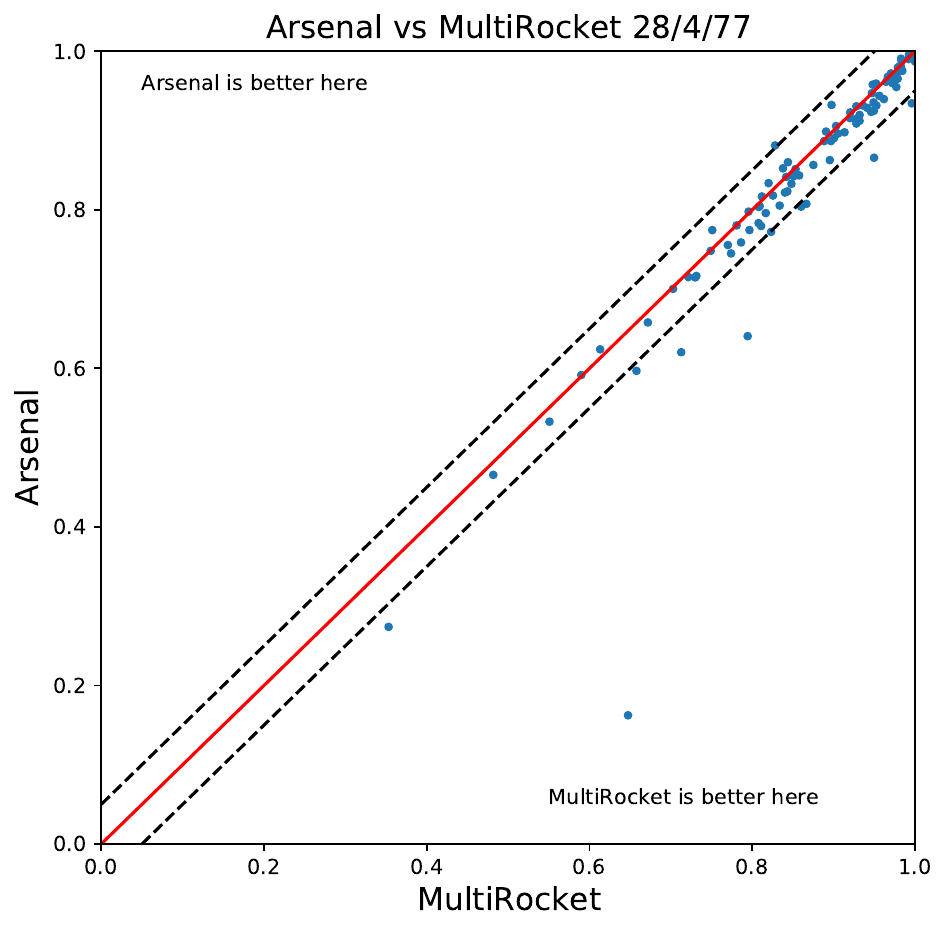}
		\caption{}
		\label{fig:arsenal vs multi}
	\end{subfigure}
	\begin{subfigure}{0.49\linewidth}
		\includegraphics[width=\linewidth]{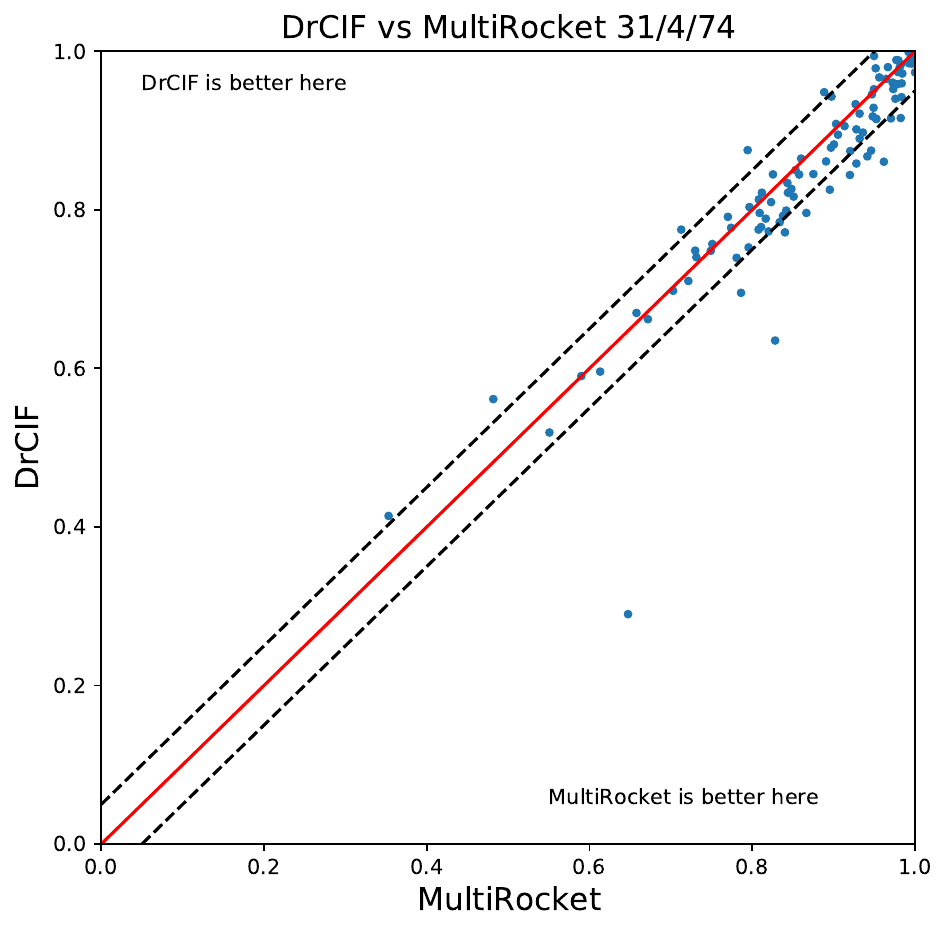}
		\caption{}
		\label{fig:drcif vs multi}
	\end{subfigure}
	\begin{subfigure}{0.49\linewidth}
		\includegraphics[width=\linewidth]{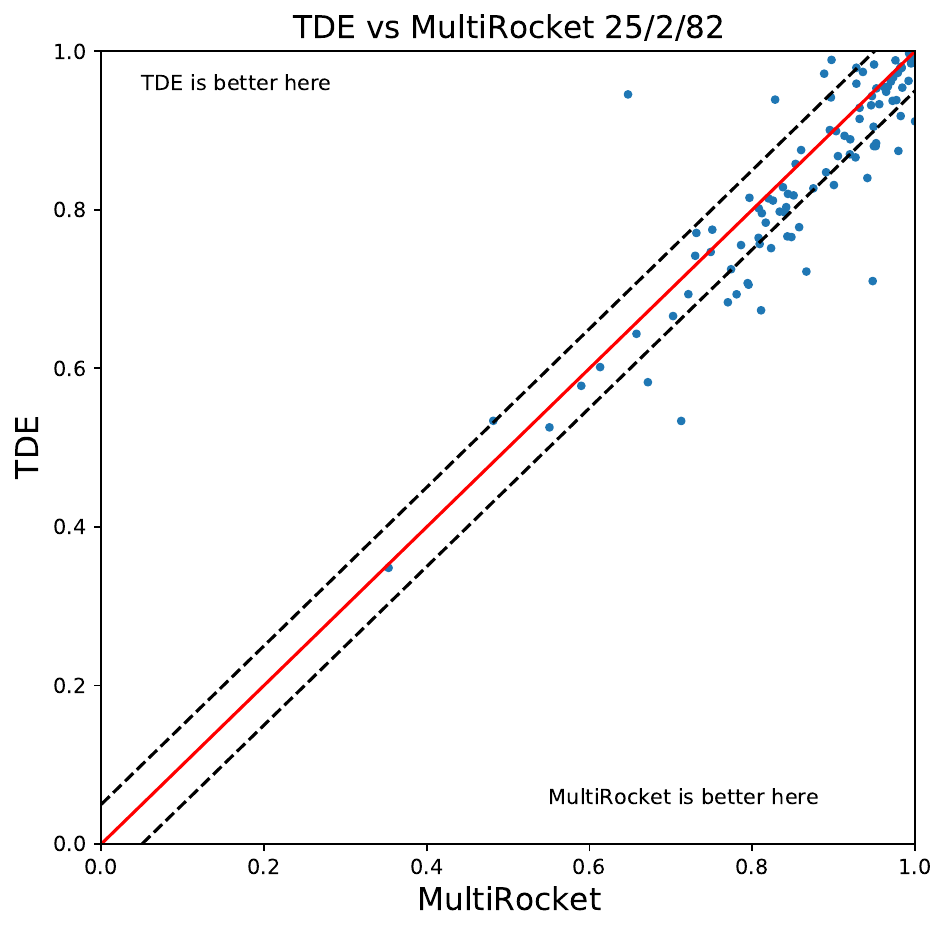}
		\caption{}
		\label{fig:tde vs multi}
	\end{subfigure}
	\caption{Pairwise accuracy comparison of \ourmethod{} against SOTA methods on 109 datasets from the UCR archive. Each point represents the average accuracy value over 30 resamples of the each dataset. The dotted lines indicate $\pm5\%$ interval on the classification accuracy.}
\end{figure}

We also compare \hctwo{} with \ourmethod{} and the existing top 3 SOTA methods.
The figures show that \ourmethod{} has similar accuracy with \hctwo{} (more datasets within the $\pm5\%$ range) than any other SOTA methods.

\begin{figure}[!h]
	\centering
	\begin{subfigure}{0.49\linewidth}
		\includegraphics[width=\linewidth]{annotated_scatter_hc2_multirocket.pdf}
		\caption{}
		\label{fig:multi vs hc2}
	\end{subfigure}
	\begin{subfigure}{0.49\linewidth}
		\includegraphics[width=\linewidth]{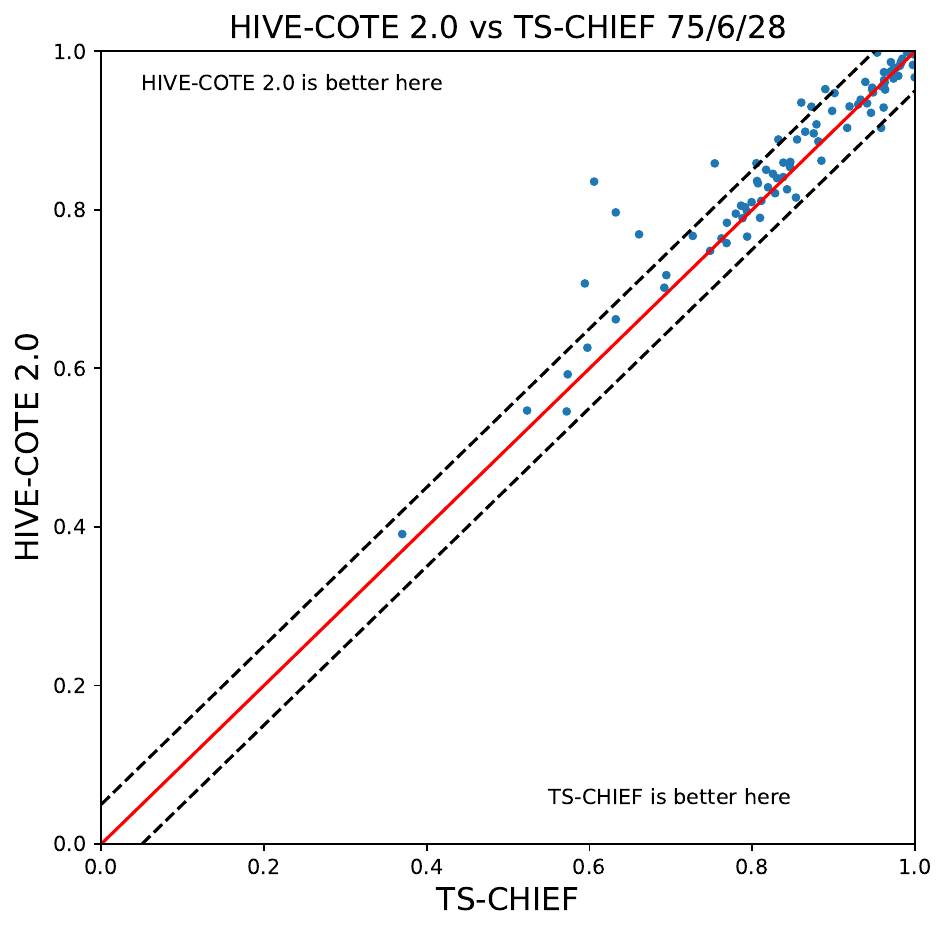}
		\caption{}
		\label{fig:tschief vs hc2}
	\end{subfigure}
	\begin{subfigure}{0.49\linewidth}
		\includegraphics[width=\linewidth]{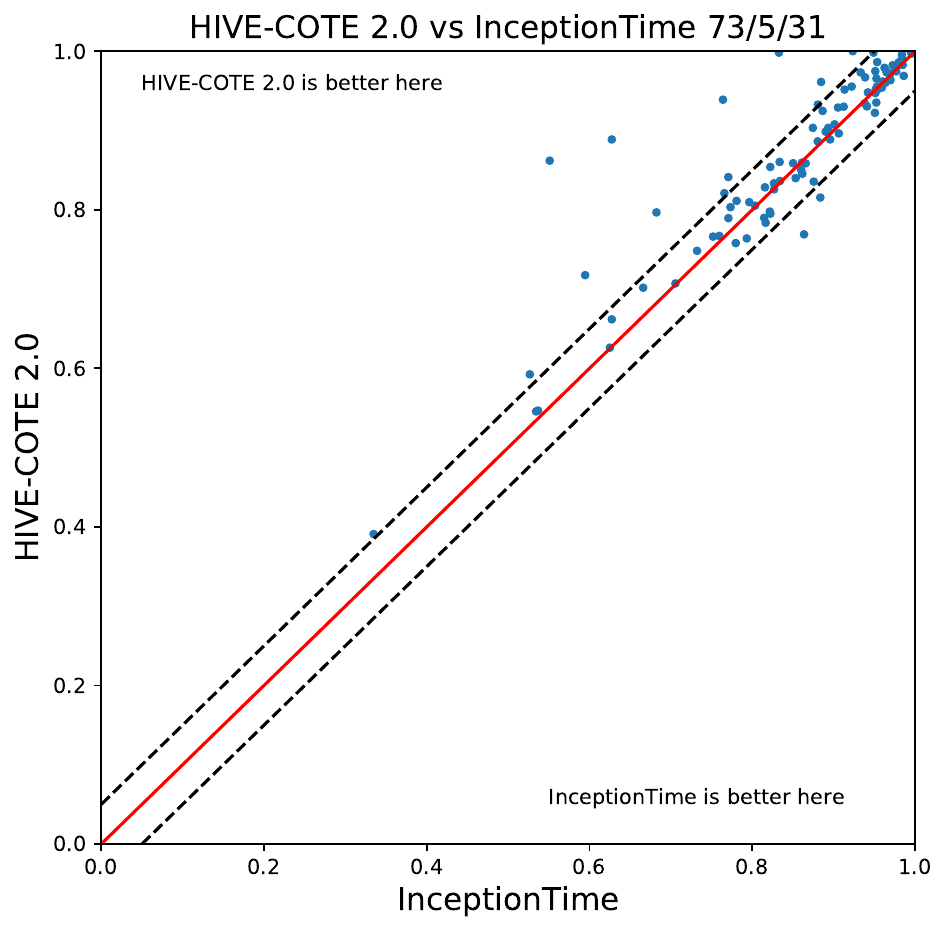}
		\caption{}
		\label{fig:inception vs hc2}
	\end{subfigure}
	\begin{subfigure}{0.49\linewidth}
		\includegraphics[width=\linewidth]{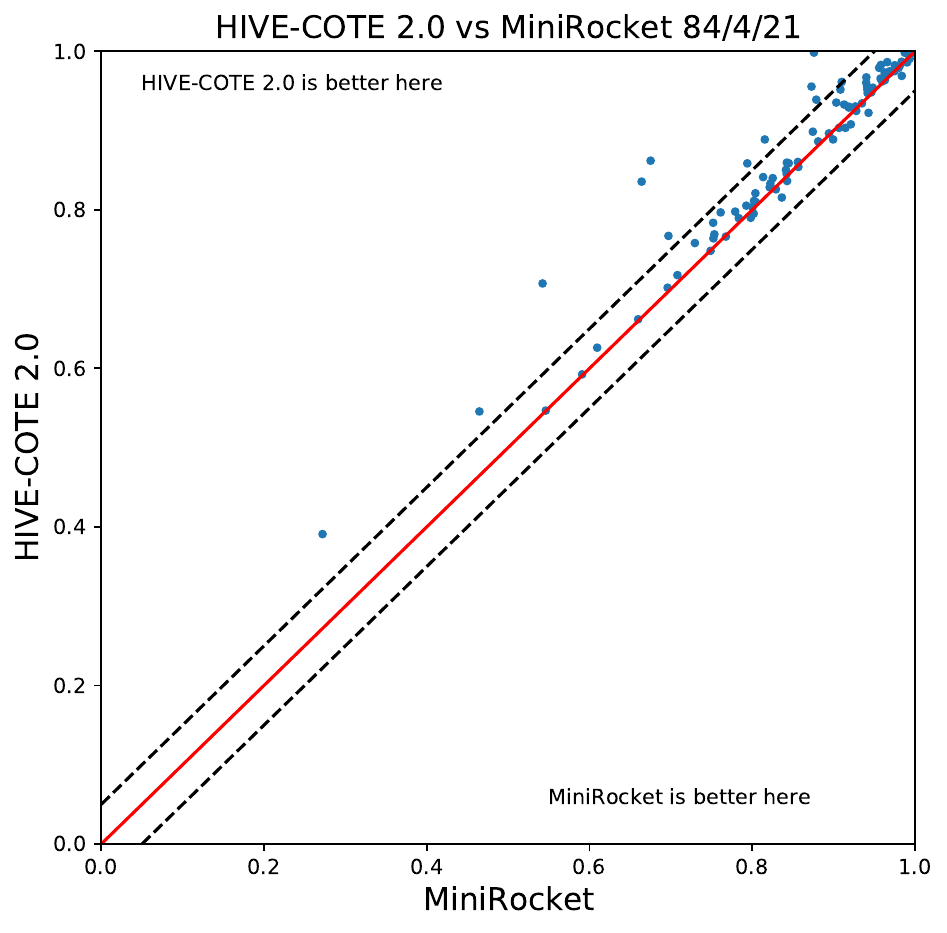}
		\caption{}
		\label{fig:mini vs hc2}
	\end{subfigure}
	\caption{Pairwise accuracy comparison of \hctwo{} against \ourmethod{} and existing top 3 other SOTA methods on 109 datasets from the UCR archive. Each point represents the average accuracy value over 30 resamples of the each dataset. The dotted lines indicate $\pm5\%$ interval on the classification accuracy.}
\end{figure}

\newpage
\section{\ourmethod{} versus \minirocket{}}
\label{app: multi vs mini}
\begin{figure}[!h]
	\centering
	\begin{subfigure}{0.49\linewidth}
		\includegraphics[width=\linewidth]{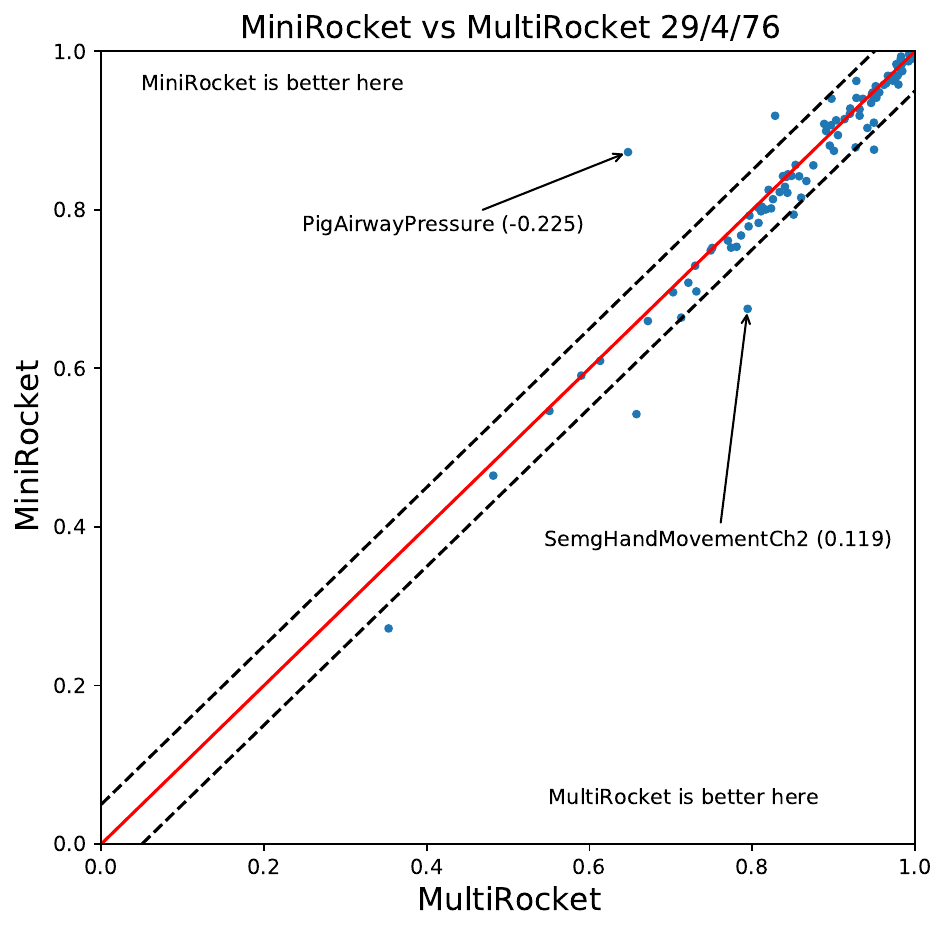}
		\caption{}
		\label{fig:vs mini}
	\end{subfigure}
	\begin{subfigure}{0.49\linewidth}
		\includegraphics[width=\linewidth]{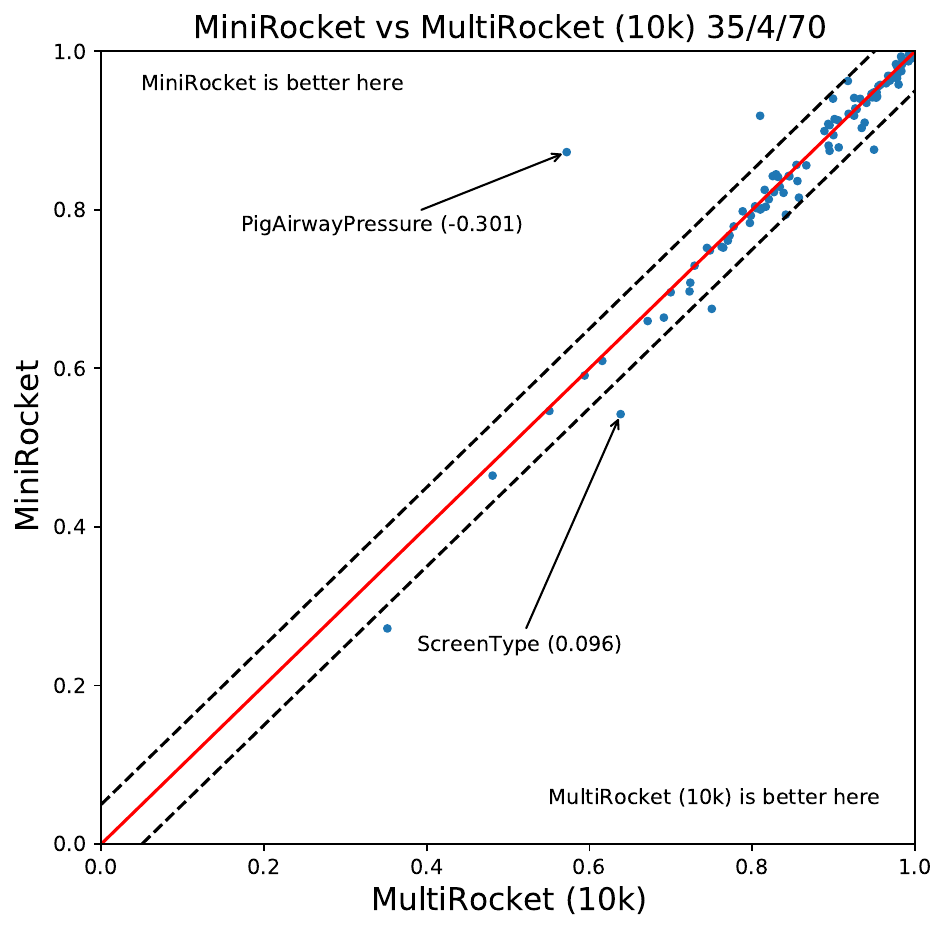}
		\caption{}
		\label{fig:vs mini 10k}
	\end{subfigure}
	\caption{Pairwise comparison of \minirocket{} versus (a) \ourmethod{} with the default 50,000 features and (b) a smaller variant of \ourmethod{} with 10,000 features on 109 datasets from the UCR archive. Each point represents the average accuracy value over 30 resamples of each dataset. The dotted lines indicate $\pm 5\%$ intervals on the classification accuracy.}
\end{figure}

This section studies the advantages and limitations of \ourmethod{} over \minirocket{}.
Figure \ref{fig:vs mini} shows the pairwise accuracy comparison of \minirocket{} and \ourmethod{} on 109 UCR datasets. 
\ourmethod{} by default generates 50,000 features, 5 times more features than \minirocket{}.
Hence, we created a smaller variant of \ourmethod{} with 10,000 features to be comparable to the default \minirocket{}, as shown in Figure \ref{fig:vs mini 10k}.

Overall, \ourmethod{} is significantly more accurate than \minirocket{}, where \ourmethod{} is consistently more accurate on 76 datasets and less accurate on 29, with 4 ties.
However, a closer look at the results indicates that most wins are within the range of $5\%$ accuracy (a phenomenon observed among the top SOTA methods, see \ref{app:pairwise comparison}) with the largest difference of 0.119 on the \texttt{SemgHandMovementCh2} dataset. 
As expected, \ourmethod{} performs the worst on the \texttt{PigAirwayPressure} dataset, with a difference of 0.225 in accuracy.
Similarly, \ourmethod{} (10k) is also significantly more accurate than \minirocket{} with 70 wins as shown in Figure \ref{fig:vs mini 10k}.

\end{document}